# Graph-augmented Segmentation for Enhanced In-Situ Geometry Reconstruction of Complex Shapes in Laser Powder Bed Fusion

Stefano Raimondo*, Matteo Bugatti, Marco Grasso
Dipartimento di Meccanica, Politecnico di Milano, Via La Masa 1, 20156 Milan, Italy
*Corresponding author. stefano.raimondo@polimi.it

**Abstract:** *In situ* geometric inspection in Laser Powder Bed Fusion of Metals (PBF-LB/M) opens new possibilities for anticipated defect detection and more efficient qualification. It relies on image segmentation to reconstruct layer-wise geometries and identify manufacturing deviations as the part is being built. However, existing approaches face a critical yet largely overlooked challenge: photometric out-of-distribution (OOD) generalization, namely the ability to maintain reconstruction accuracy when image appearance differs from the conditions represented during model training. This study represents the first exploration of a graph-augmented U-Net architecture for *in situ* geometric inspection in PBF-LB/M. The hybrid network integrates a Graph Neural Network (GNN) within the bottleneck to enhance reconstruction robustness under photometric distribution shifts. It is evaluated against conventional U-Net and active contour segmentation methods using two experimental case studies involving different lattice geometries, materials, and manufacturing setups, with X-ray computed tomography providing reference geometries. Generalization is assessed across different training and inference conditions and under controlled image perturbations. Results show that the UNet-GNN achieves higher reconstruction accuracy and consistency than the benchmark methods, with more pronounced improvements when training and inference conditions differ. Its advantages are confirmed across both case studies and under controlled photometric perturbations, while preliminary anomaly detection experiments indicate that improved robustness does not necessarily compromise sensitivity to geometric deviations. These findings highlight graph augmentation as a promising strategy for more reliable *in situ* geometric inspection under heterogeneous manufacturing conditions.

## I. Introduction

Additive Manufacturing (AM) has reached increasing technological maturity across different industrial sectors, including aerospace, energy, biomedical, and creative industries. This progress has been supported by continuous improvements in process quality and system reliability, the integration of AM technologies into advanced production lines, increasing levels of automation, and developments in material processing capabilities. Nevertheless, industrial qualification and certification remain major bottlenecks for the large-scale deployment of AM technologies [1], [2]. Current qualification frameworks are still largely inherited from conventional manufacturing routes and rely heavily on expensive inspection methodologies, destructive testing, and conservative safety margins. These limitations become particularly critical for innovative AM applications involving complex and unconventional geometries, where traditional inspection methods may be technically challenging or economically impractical. As a result, the gap between manufacturing capabilities and qualification efficiency continues to limit the exploitation of metal AM in industrial environments [2], [3].

To address this challenge, *in situ* monitoring and inspection approaches leveraging layer-wise data acquired during fabrication have emerged as enabling technologies for more efficient and reliable quality assurance [3]–[9]. In metal laser powder bed fusion (PBF-LB/M), optical images of the powder bed can be acquired after laser exposure to extract spatially resolved information about the solidified regions. Image segmentation can then be employed to reconstruct the geometry of each manufactured layer and compare it with the corresponding nominal geometry derived from the digital part model [7]–[9]. This enables the identification of geometric deviations during fabrication, potentially allowing non-conformities to be detected before part completion and supporting a transition from predominantly post-process inspection towards *in situ* quality verification [8], [9].

A growing body of research has investigated *in situ* and layerwise geometric reconstruction and inspection in PBF-LB/M [10]–[21]. A widely explored family of approaches is based on active contour segmentation [11], [12], [14], [18]. These methods iteratively refine an initial boundary through shrinkage and expansion operations driven by the minimization of an energy functional, often within a level-set framework [22]. In PBF-LB/M, the contour can be initialized using the nominal CAD geometry, thereby incorporating prior knowledge of the expected shape. However, the dependence on nominal contour initialization may introduce a conservative bias towards the expected geometry, potentially limiting the ability to reconstruct actual deviations. The iterative nature of the optimization also introduces computational requirements that can constrain real-time applicability, particularly when processing high-resolution powder bed images at industrial scale.

More recently, convolutional neural network (CNN) models, particularly U-Net architectures, have demonstrated promising performance and computational efficiency for *in situ* geometric reconstruction in PBF-LB/M [23]–[25]. By learning hierarchical image representations directly from annotated training data, these architectures can capture complex patterns and provide accurate pixel-wise segmentation without requiring explicit initialization from nominal contours. Nevertheless, learned representations may remain sensitive to local appearance patterns and their statistical characteristics. When photometric properties deviate from those sufficiently represented during training, the resulting distribution shift may reduce segmentation accuracy and consistency [26]. This is known as photometric out-of-distribution (OOD) generalization issue, namely the difficulty of a model to maintain reliable performance when inference data exhibit statistical or structural characteristics that are not adequately represented in the training distribution [26], [27].

In the context of powder bed image segmentation, the OOD generalization challenge can arise from at least two distinct sources. The first concerns variations in the photometric properties of the acquired images, such as changes in illumination, local contrast, surface reflectivity, and pixel intensity distributions. These variations may either alter the appearance of otherwise identical geometric or masking actual changes in the manufactured geometry. The second concerns geometric OOD conditions, where the shape of the solidified regions differs from the geometries represented during training, because of dimensional or geometric deviations. Although both aspects are relevant to *in situ* inspection, they pose different challenges: photometric OOD generalization requires maintaining reconstruction consistency despite changes in image appearance, whereas geometric OOD generalization requires preserving sensitivity to actual geometric features that may not have been encountered during training. The present study focuses primarily on the former challenge.

Photometric OOD generalization is particularly relevant in PBF-LB/M as variations may occur within the same layer, due to illumination inhomogeneities and camera orientation effects, or between successive layers [18]. Layer-wise changes can arise, for example, from the rotation of the laser scanning direction, which influences the orientation of surface waviness in the solidified regions and consequently the reflection of light towards the camera [12]–[18]. Depending on the relative orientation of the scanning tracks, illumination source, and image acquisition settings, the resulting images may range between so-called dark-field and bright-field configurations [12]. These effects introduce changes in pixel intensity distributions and local contrast that are not necessarily associated with differences in the underlying manufactured geometry.

Fig. 1 illustrates the severity of such variability through three representative powder bed images acquired during the PBF-LB/M of copies of the same lattice structure. All images were acquired using a fixed camera and identical acquisition settings. The examples correspond either to the same location of the build area at different layers (left and central panels) or to different locations within the same layer (central and right panels). The associated pixel intensity histograms highlight how spatial and layer-wise changes in imaging conditions can substantially modify the photometric characteristics of otherwise comparable geometric features.

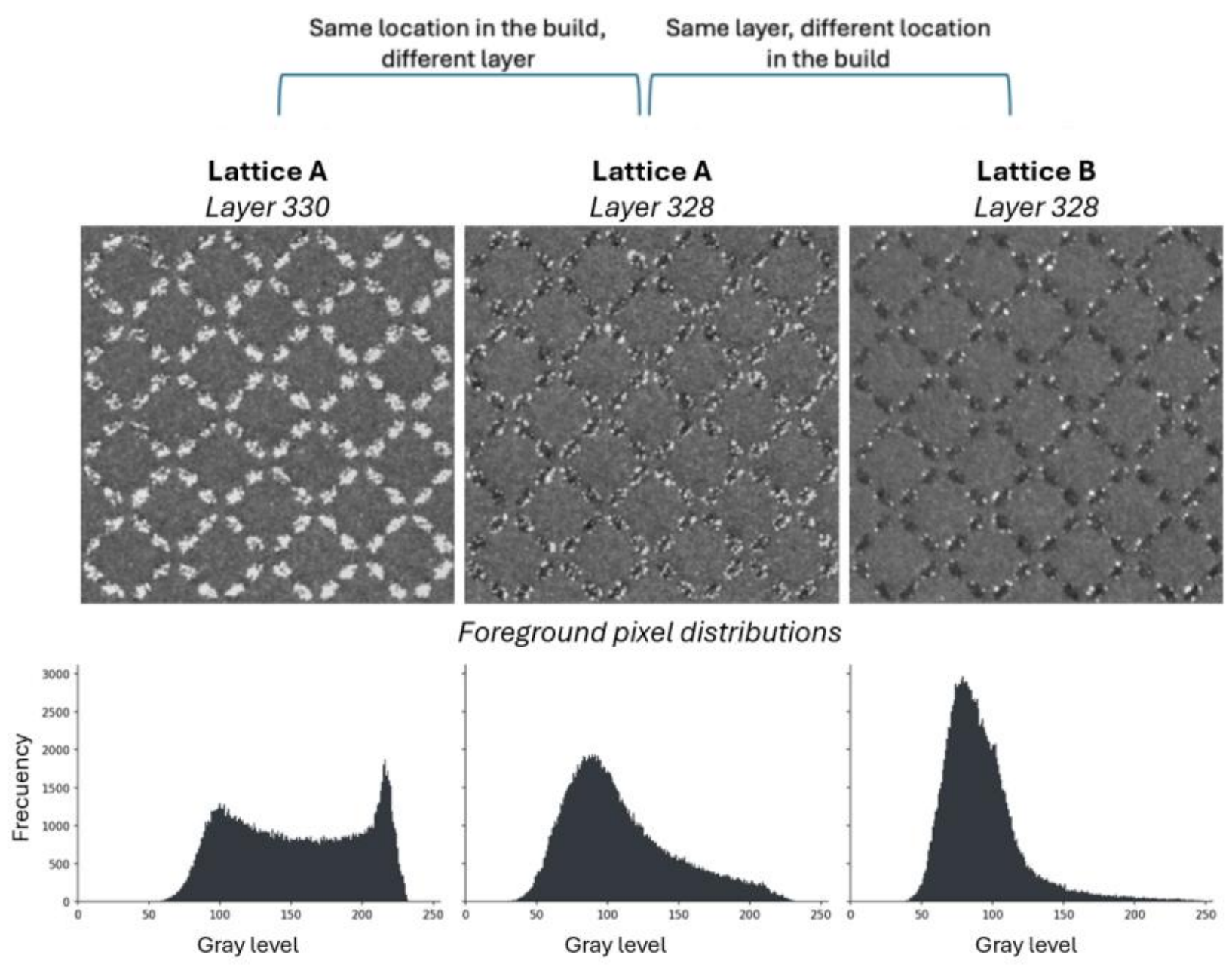

**Fig. 1.** Examples of powder bed images acquired during the PBF-LB/M of copies of the same lattice structure in different build locations and across different layers, along with their corresponding pixel intensity histograms.

A reliable and trustworthy inspection thus requires sufficient invariance to nuisance photometric variability while preserving sensitivity to meaningful geometrical features.

The most common approaches to tackle the photometric OOD generalization issue consist of increasing the amount of training data to cover a wider natural variability or augmenting the training data through controlled transformations of the original images (e.g., variations in brightness, contrast, gamma correction, signal-to-noise ratio, etc.) [26], [27]. However, both strategies have inherent limitations. Expanding the training dataset requires additional experimental campaigns and ground-truth annotations, which can be costly and time-consuming. Data augmentation, on the other hand, may not adequately reproduce the complexity and full range of photometric variations encountered under real manufacturing conditions. Moreover, neither strategy inherently addresses the dependence of learned feature representations on specific photometric patterns, potentially limiting generalization to previously unseen imaging conditions.

The alternative direction explored in this study consists of investigating a network architecture capable of representing the underlying geometry through features that are less dependent on specific local intensity patterns. In this perspective, explicitly modeling relationships between spatial regions may provide complementary information to conventional convolutional feature extraction, supporting more consistent reconstruction when image appearance changes. Following this route, we propose tackling the OOD generalization issue for *in situ* geometry reconstruction in PBF-LB/M by means of a graph-augmented U-Net architecture, which combines the potential of consolidated U-Net architectures with a Graph Neural Network (GNN).

GNNs represent data through interconnected nodes, whose features are iteratively updated by exchanging and aggregating information from neighboring nodes through message-passing operations [28], [29]. Unlike conventional convolutions, which operate over predefined spatial neighborhoods, graph-based operations can model relationships between regions that are not necessarily spatially adjacent, potentially capturing long-range dependencies and structural information. The hypothesis explored in this study is that such capability may support enhanced geometric reconstructions by reducing reliance on local photometric characteristics.

Hybrid CNN–GNN architectures have attracted growing attention in recent years, particularly in medical image analysis [30]–[37]. In these applications, graph-based representations have been primarily exploited to improve segmentation accuracy when dealing with complex geometries, ambiguous boundaries, or image patterns that are difficult to resolve through alternative methods. However, their potential to enhance photometric OOD generalization remains comparatively unexplored, particularly in the context of *in situ* inspection in PBF-LB/M.

To address this gap, the present study investigates a hybrid UNet–GNN architecture, in which a GNN is embedded within the bottleneck of a conventional U-Net to enrich convolutional features with relational information. The specific objective is to improve the robustness and consistency of geometric reconstruction under photometric distribution shifts, while preserving reconstruction accuracy and limiting the amount of training data required.

The motivating application considered in this study is the *in situ* geometric reconstruction of lattice structures. These complex periodic architectures derive their functional properties primarily from their geometry rather than from the constituent material alone [38]. Consequently, local geometric inaccuracies, such as strut thinning, missing nodes, and dimensional distortions, can significantly affect their mechanical, thermal, and functional performance [39]–[44]. Their qualification represents a particularly demanding industrial challenge, as conventional non-destructive inspection methods, such as X-ray computed tomography (CT), are costly and time-consuming and may become impractical for large or dense components. The development of *in situ* geometric inspection methods is therefore especially relevant for these structures [45].

The present study builds on previous investigations of *in situ* inspection of lattice structures [46], [47], although the proposed methodology is not restricted to a specific geometry and can potentially be extended to other classes of manufactured components.

The UNet-GNN architecture is experimentally evaluated using powder bed images acquired during the PBF-LB/M of lattice structures of different types and materials, manufactured with different machines and at different locations within the build area. Its performance is benchmarked against two representative segmentation approaches, namely active contours and a conventional U-Net. Quantitative validation is performed using ground-truth geometries derived from X-ray CT scans of the as-built specimens. To investigate the influence of variability between training and inference conditions, different training configurations are considered, including separate and joint training strategies involving specimens exposed to different imaging conditions. In addition to real data variations, OOD generalization is systematically assessed under controlled image perturbations, including additive Gaussian noise, gamma-based intensity transformations, and progressive pixelation at increasing severity levels. These experiments provide complementary evidence on the ability of the segmentation method to maintain accurate geometric reconstruction when image properties differ from those represented during model development.

Finally, artificially introduced geometric deviations are considered to assess whether improved photometric OOD generalization can be achieved without compromising the sensitivity of the segmentation-based inspection framework to actual departures from nominal geometry.

The main contributions of this work can be summarized as follows:

- It presents the first application of graph-augmented segmentation to *in situ* geometric reconstruction in PBF-LB/M.
- It addresses photometric OOD generalization as an overlooked challenge in PBF-LB/M inspection, highlighting the potential of graph-augmented U-Net architectures to improve reconstruction accuracy and consistency under photometric distribution shifts.
- The proposed approach is evaluated across different geometries, materials, and printing configurations, complemented by controlled image perturbations and comparisons with established segmentation methods.

Overall, the findings contribute to understanding how graph-based feature representations can support more reliable and accurate *in situ* geometric reconstruction under realistic manufacturing variability, providing a foundation for the development of segmentation-based inspection systems capable of operating under heterogeneous industrial imaging conditions.

The remainder of the paper is organized as follows. Section II presents the adopted methodology and benchmarking approaches. Section III describes the experimental case study and numerical settings. Section IV discusses the results and performance comparisons. Section V summarizes the main conclusions.

## II. Methodology

### *2.1 In situ geometry reconstruction and benchmark methods*

In PBF-LB/M, *in situ* geometry inspection based on post-exposure powder bed images is performed by acquiring a top-view image of each layer after laser scanning and before powder recoating [7]–[9], [48]. At this stage, the solidified regions exhibit distinct optical characteristics compared to the surrounding loose powder, mainly due to differences in surface morphology and reflectivity. Acquired images are then pre-processed to compensate for lens distortions and camera perspective, which is typically done in a calibration stage. Subsequently, image segmentation is applied to reconstruct the contour of the solidified region.

The primary objective of *in situ* geometry reconstruction is to quantify, on a layer-by-layer basis, both local and global deviations from the nominal geometry. The reconstructed geometry therefore provides the basis for *in situ* identification of anomalous geometric distortions through comparison with the corresponding nominal slice derived from the CAD model. This study primarily focuses on the photometric OOD generalization faced during the geometry reconstruction step, but the use of the UNet-GNN methodology for automated anomaly detection is presented and discussed in sub-section 2.3.

The two segmentation approaches most widely adopted in the PBF-LB/M literature for *in situ* geometry reconstruction are active contours [14], [18], and U-Nets [24],[25], and they are used as benchmarks in this study.

The active contour methodology is an iterative segmentation algorithm. It evolves an initial contour by minimizing an energy functional, progressively refining the boundary of the foreground region through successive iterations [49]-[51]. Such iterations involve both internal and external “forces”. The internal term imposes smoothness constraints to prevent irregular or fragmented shapes. The external term is derived from image features and attracts the contour toward meaningful structures in the image. In level-set formulations, the contour is represented implicitly as the zero level of a higher-dimensional function. The evolution is governed by the solution of partial differential equations obtained from the Euler–Lagrange minimization of the total energy functional. Through iterative updates, the contour converges to a configuration that balances geometric regularity with adherence to background-foreground separation features, resulting in the delineation of reconstructed boundaries.

Over the years, numerous variants have been proposed, evolving from edge-based to region-based formulations and hybrid models [49]–[51]. Region-based approaches rely on statistical differences between pixel intensity maps inside and outside the evolving contour. In contrast, edge-based models drive contour evolution through image gradient information, attracting the curve toward sharp local intensity transitions that correspond to object boundaries. Recent state-of-the-art formulations increasingly combine region- and edge-based terms within unified energy functionals, aiming to leverage complementary information from both global intensity statistics and local gradient cues [14], [50]. Active contours attracted particular interest in seminal studies on *in situ* monitoring and inspection in PBF-LB/M due to the possibility of initializing the contour directly from the nominal CAD geometry [10], [12], [14], [18]. This initialization strategy can represent a significant advantage, as it constrains the evolution process around an expected boundary and improves convergence stability, especially in the presence of noise or weak edges. However, it may also constitute a limitation. If the regularization terms dominate the energy functional, the contour may remain overly biased toward the nominal geometry, reducing its sensitivity to actual geometric deviations, which are precisely the target of *in situ* inspection. Conversely, relaxing the regularization to enhance deviation detectability may increase susceptibility to noise and spurious image artifacts. Achieving an appropriate balance between fidelity to the nominal initialization and the ability to reconstruct real geometric discrepancies therefore represents a critical methodological challenge [51]. A further limitation concerns computational cost, which is inflated by the iterative nature of the method.

In contrast to the active contour methodology, U-Net is a fully convolutional neural network architecture that does not require an explicit contour initialization step. Instead, it relies on supervised training to learn the mapping between input images and segmentation masks, achieving the desired level of accuracy through data-driven optimization [52]–[54]. Originally introduced for biomedical image segmentation, U-Net and its variants have since become a reference architecture in a wide range of computer vision applications due to its ability to combine precise localization with robust semantic understanding [53].

The network employs a symmetric encoder–decoder (contracting–expanding) structure. The encoder path progressively reduces the spatial resolution of the input through a sequence of convolutional layers followed by max-pooling operations. This hierarchical feature extraction allows the network to capture increasingly abstract and contextual representations of the input image at multiple spatial scales. As depth increases, the receptive field grows, enabling the model to incorporate global contextual information that is crucial for accurate segmentation.

The decoder path mirrors the encoder structure and gradually restores the spatial resolution of the feature maps using up-convolution (transposed convolution) layers. At each decoding stage, the up-sampled feature maps are concatenated with corresponding high-resolution features from the encoder through skip connections. These skip connections play a critical role by directly transferring fine-grained spatial information that would otherwise be lost during down-sampling [55]–[57]. As a result, the decoder can effectively combine low-level spatial details with high-level semantic features, leading to improved boundary identification accuracy.

U-Nets have been applied to powder bed images in PBF-LB/M by various authors [24],[25]. However, although deeper layers provide broader contextual information, the final segmentation remains locally resolved at the pixel level. This characteristic may limit the OOD generalization in scenarios where image properties vary from layer to layer and from build to build, as commonly observed in industrial PBF-LB/M applications. In such cases, local intensity fluctuations may lead to inconsistent feature representations and reduced generalization capability, potentially yielding segmentation inaccuracies when the model is exposed to domain shifts not sufficiently represented in the training data.

### *2.2 Graph-augmented U-Net segmentation*

Active contour models and U-Net architectures operate either through local boundary evolution mechanisms or dense pixel-wise feature learning. In applications like *in situ* geometry inspection, however, segmentation is not merely a matter of boundary detection but also of preserving structural coherence and connectivity. Purely local or pixel-driven approaches may fail to enforce global structural consistency, potentially leading to poor reconstruction accuracies under challenging imaging conditions.

Graph-based segmentation has recently emerged as a promising strategy to address several of these limitations, but in application domains different from AM [28], [37], [58]. GNNs introduce an additional level of contextual information by explicitly connecting and exchanging information between different image regions, including regions that are spatially distant but exhibit similar learned features [28], [29]. As a result, the representation of a given region is influenced not only by its local appearance but also by its relationships with other regions. This may help the network maintain more consistent feature representations when local photometric characteristics change, reducing its dependence on specific intensity patterns observed during training.

Different variants of possible CNN–GNN architectures have been proposed so far across different application domains, including medical image analysis and autonomous driving [28], [30]–[37]. In this study, we specifically consider a hybrid architecture where the U-Net bottleneck is enhanced by a GNN. This design, hereafter referred to as UNet-GNN, leverages the convolutional feature extraction capabilities of the U-Net while introducing a graph-based mechanism to capture complex global geometrical interdependencies. Specifically, intermediate feature maps are reinterpreted as graph-structured representations, where nodes encode spatially aggregated features, whereas edges model relational dependencies between neighbouring or structurally connected regions. Through message-passing operations, the GNN propagates information across these nodes, enabling the network to learn not only local appearance cues but also higher-order connectivity patterns and structural constraints.

The model consists of three main components, as illustrated in Fig. 2:

*Encoder:* this is the component responsible for extracting hierarchical feature representations through a series of convolutional blocks. These layers progressively capture low-level to high-level spatial features.
*GNN bottleneck:* it is designed to learn global geometric context and relational dependencies by leveraging GNN layers. At this stage, feature representations are modeled as graph-structured data, enabling the network to capture long-range interactions and structural information.
*Decoder:* in charge of progressively reconstructing the spatial resolution and generating the final segmentation map, combining contextual information from the bottleneck with fine-grained spatial details.

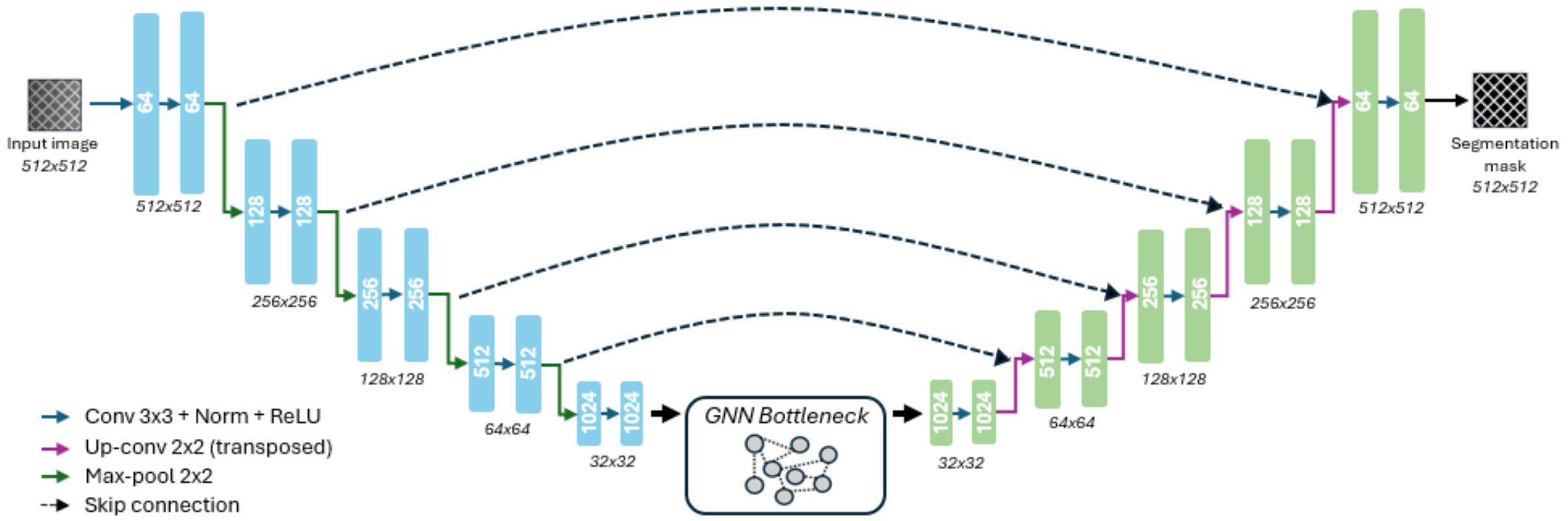

**Fig. 2.** Hybrid CNN-GNN model architecture: the network is composed of three main modules: an encoder, a GNN bottleneck, and a decoder.

Let $L$ be the number of convolutional blocks, such that the $L$-th block is the deeper one, and let $H \times W$ be the size (in pixel) of the input images (note that in this study we refer to greyscale images, but the method can be directly generalized to 3-channel RGB images). Then, the deepest encoder feature map is referred to as $F_L \in \mathbb{R}^{H_L \times W_L \times C_L}$, where $C_L$ is the number of feature channels at the $L$-th level.
The graph is formulated as $G = (V, E)$, where $V$ denotes nodes and $E$ edges. Each spatial location $(x, y)$ in the feature map $F_L$ becomes a node, i.e., $v_{x,y} \in V$, thus the number of nodes is equal to $H_L \times W_L$, and the initial node feature, denoted by $h^0_{x,y}$, coincides with $F_L(x, y)$. Edges, instead, are created using the $k$-nearest neighbours (k-NN) formulation, such that each node $v_{x,y}$ connects to the $k$ closest nodes:

$$E = \{(v_{x,y}, v_{x\prime,y\prime}) | v_{x',y'} \in kNN(v_{x,y})\}.$$

where $kNN(v_{x,y})$ denotes the set of $k$ nearest neighbours of node $v_{x,y}$.
Node features in the $t$-th layer of the GNN, namely $h^t_{x,y}$, are updated by aggregating and weighting information from the neighbours of the node in the $(t$-1)-th layer, by means of a rectified linear unit (ReLU) activation function $\sigma$, with weight $W^t$ and bias $b^t$. Thus, the graph convolution in the $t$-th layer is given by:

$$h_{x,y}^{t} = \sigma\left(\sum\nolimits_{j\in kNN(v_{x,y})} W^{t}\left(h_{j}^{t-1}\right) + b^{t}\right)$$

In our implementation, two GNN message-passing layers are stacked within the GNN bottleneck (i.e., t = 1, 2). The graph topology is computed once based on the initial node features and kept fixed across the message-passing layers. The parameter $k$ (number of neighbours for the construction of the graph) was set to $k = 8$. Increasing $k$ expands the set of directly connected neighbors, whereas increasing the number of GNN layers allows information to propagate through successive graph connections. Both mechanisms can enrich contextual representations, although excessive aggregation may reduce sensitivity to localized feature differences through over-smoothing and increase computational requirements. A value $k = 8$ provides a compromise between preserving local geometric relationships and enabling sufficient information exchange across the graph. Smaller values of $k$ may lead to sparse graph connectivity and limited contextual aggregation, whereas excessively large values increase the risk of over-smoothing node representations, potentially reducing sensitivity to localized geometric deviations.

Distances among neighbours are expressed in terms of Euclidean distance in the feature space, which at the bottleneck is of size $H_L = W_L = 32$ and $C_L = 1064$.

Table I summarizes the main differences between the standard U-Net architecture and the UNet-GNN.

**TABLE I** - Structural comparison between the standard U-Net architecture and UNet-GNN

| Component | U-Net | UNet-GNN |
|---|---|---|
| Encoder | Convolutional blocks (DoubleConv + MaxPooling): L=4blocks, base channels = 64 | |
| Feature Extraction | Hierarchical multi-scale CNN features | |
| Bottleneck module | Pure convolutional feature maps | Graph-based bottleneck applied to deepest feature map |
| Graph Construction | Not present | Graph with k-NN edges built on bottleneck feature embeddings |
| Graph Convolution | Not present | Two stacked message-passing layers |
| Decoder | Transposed convolutions + skip connections | |
| Skip Connections | Encoder–decoder concatenation | |
| Output Layer | 1×1 convolution | |

*2.3 In situ detection of geometrical deviations*

To evaluate whether improved OOD generalization to undesired sources of variability in powder-bed image segmentation is achieved without compromising the ability to identify actual deviations from the nominal geometry, a control chart inspired by the methodology proposed in [14] was applied in this study. The control chart was originally proposed in combination with an active contour segmentation methodology, but it is here generalized to the UNet-GNN framework. The underlying principle consists of evaluating the deviation of the reconstructed contour in one layer from the corresponding nominal contour in the same layer.

Let $d_i(j)$ be the Euclidean distance between the j-th pixel of the reconstructed contour and the closest point belonging to the nominal contour for the i-th layer, evaluated on a signed deviation map (positive outside the nominal region, negative inside). The minimum and maximum deviations from the nominal contour in the i-th layer are computed as:

$$d_i^{max} = max_j d_i(j),\ d_i^{min} = min_j d_i(j)$$

and the synthetic monitoring index is the maximum absolute value of the two extreme deviations along the reconstructed contour:

$$d_i^{minmax} = \max\{|d_i^{max}|, |d_i^{min}|\}$$

A negative $d_i^{min}$ corresponds to a portion of the reconstructed slice that is missing with respect to the nominal geometry, whereas a positive $d_i^{max}$ corresponds to an excess of material with respect to the nominal shape.

In its original formulation, $d_i(j)$ is sampled exclusively along the reconstructed contour [14]. This implicitly assumes that the reconstructed and nominal contours are topologically comparable, i.e., that a reconstructed boundary exists in the neighborhood of every relevant nominal feature. This assumption breaks down when an entire geometrical feature (e.g., a strut section) is not segmented at all, as can occur under the induced high severity anomaly: in this case, no reconstructed contour pixel exists in the vicinity of the missing feature, and the corresponding deviation is never sampled, causing the defect to go undetected even though it is arbitrarily large.

To address this limitation, the deviation index is extended to a bidirectional formulation, in a way inspired by the work [59] that was applied to point-cloud data. In addition to $d_i(j)$, let $d_i^{nom}(k)$ be the Euclidean distance between the k-th pixel of the nominal contour and the closest point belonging to the reconstructed contour for the i-th layer. The set of sampled deviations is then extended as:

$$D_i = \{d_i(j)\}_j\ U\ \{-d_i^{nom}(k)\}_j$$

where the sign of $d_i^{nom}(k)$ is reversed so that it is consistent with the missing-material convention of $d_i^{min}$. The extreme deviations and the synthetic index are then computed over the extended set:

$$d_i^{max} = \max(D_i),\ d_i^{min} = \min(D_i),$$

$$d_i^{minmax} = \max\{|d_i^{max}|, |d_i^{min}|\}$$

This bidirectional formulation ensures that a geometrical deviation consisting of a local lack of material in the solidified layer is still captured through the nominal-to-reconstructed direction, since its nominal contour pixels are, by construction, far from any reconstructed pixel, yielding a large negative contribution to $d_i^{min}$.

The evolution of the statistic $d_i^{minmax}$ throughout the build is monitored using a one-sided adaptive control chart. The upper control limit (UCL) is estimated assuming that $d_i^{minmax}$ follows an extreme-value Gumbel distribution, whose parameters are initialized from an initial set of $n$ in-control layers, i.e., layers free of anomalous deviations, by maximum likelihood estimation. More precisely, the UCL is defined as the $1-\alpha$ percentile of the Gumbel distribution in the $n$ in-control layers, where $\alpha$ is the target false alarm rate of the control chart.

A layer is classified as anomalous whenever the corresponding $d_i^{minmax}$ exceeds the current UCL.

As introduced in [14], the monitoring scheme adopts a moving-window updating strategy to account for the layer-dependent evolution of the nominal geometry. Provided that no alarm is raised, the distribution parameters and the corresponding control limit are recursively re-estimated including the last monitored layer, allowing the chart to adapt to the natural variability associated with changing slice geometries while preserving sensitivity to abrupt process-induced deviations.

As a final remark, it is worth noticing that a fine alignment between the *in situ* reconstruction and the nominal geometry is needed. Such alignment can be carried out during a camera calibration stage, e.g., employing landmark registration methods. For a review of these methods, the reader is referred to [12]. For additional details about the control charting scheme, instead, the reader is referred to [14].

## III. Experimental and numerical settings

### *3.1 Real case studies*

The real case studies adopted here are based on two different experimentations previously introduced respectively in [12] and in [60]. The aim is to test the proposed approach considering two different geometries manufactured with different materials, different PBF-LB/M machines, and using different image acquisition setups.

In the real case study 1, copies of trabecular lattice structures were manufactured using a TruePrint 3000 PBF-LB/M system by Trumpf. They were produced using a gas atomized maraging steel powder with a powder size distribution in the range 15 - 45 μm (D10 – D90). The printed lattice structures belong to the trabecular lattice family, and each specimen was composed of 64 equal rhombic cells (Fig. 3a – left panel) placed in a $4 \times 4 \times 4$ grid and with a strut diameter equal to 1.5 mm. The PBF-LB/M machine was equipped with a Basler acA3800–14uc USB 3.0 camera placed above a viewport on the top of the build chamber (Fig. 3a, central panel). The field of view covered the whole build area, namely a circular region with diameter of 300 mm, providing a spatial resolution after perspective correction of about 100 μm/pixel. A light source was available inside the build chamber, inclined at about 60° with respect to the building plate and placed on the right-hand side on the chamber.

In the real case study 2, copies of gyroid lattice structures were manufactured using a 3D-NT LLA150 PBF-LB/M system by Prima Additive featuring a fiber laser source (AFX-1000, nLIGHT). They were produced using a gas atomized AlSi10Mg powder with a powder size distribution in the range 15 - 50 μm (D10 – D90). The printed lattice structures belong to the triply periodic minimal surface (TPMS) lattice family, and each specimen was composed of 64 equal TPMS cells (Fig. 3b – left panel) placed in a $4 \times 4 \times 4$ grid and with a wall thickness of 0.8 mm. The PBF-LB/M machine was equipped with an IDS UI-5490SE-M-GL powder bed camera, featuring a 10.55 Mpix sensor, which enabled a spatial resolution of about 57 μm/pixel within a field of view that embraced the whole build area. A directional light source was placed beside the powder bed camera, exploiting the second viewport on the ceiling of the build chamber (Fig. 3b – central panel).

Fig. 3 shows the *in situ* inspection and lighting configuration in the two case studies, together with the specimen location within the build area and the shape of the unit lattice structure manufactured in each case study. Both these two experimental conditions were selected as deemed representative of PBF-LB/M industrial settings.

Each specimen in both case studies, regardless of the specific type of lattice configuration, can be represented as three-dimensional stacks of $N$ unit cells with fixed shape and size. The unit cells are arranged side by side in a regular grid along the $x$ and $y$ directions and stacked along the $z$ direction, which coincides with the build direction. All unit cells are assumed to share identical bounding box dimensions and to consist of the same number of layers, which is consistent with the most common configurations of such complex structures.

In both case studies, the overall dimension of each manufactured lattice specimen was 40 × 40 × 40 mm, with unit cell size $l$ = 10 mm. Table II summarizes the main process parameters adopted to produce the specimens with the two different experimental settings.

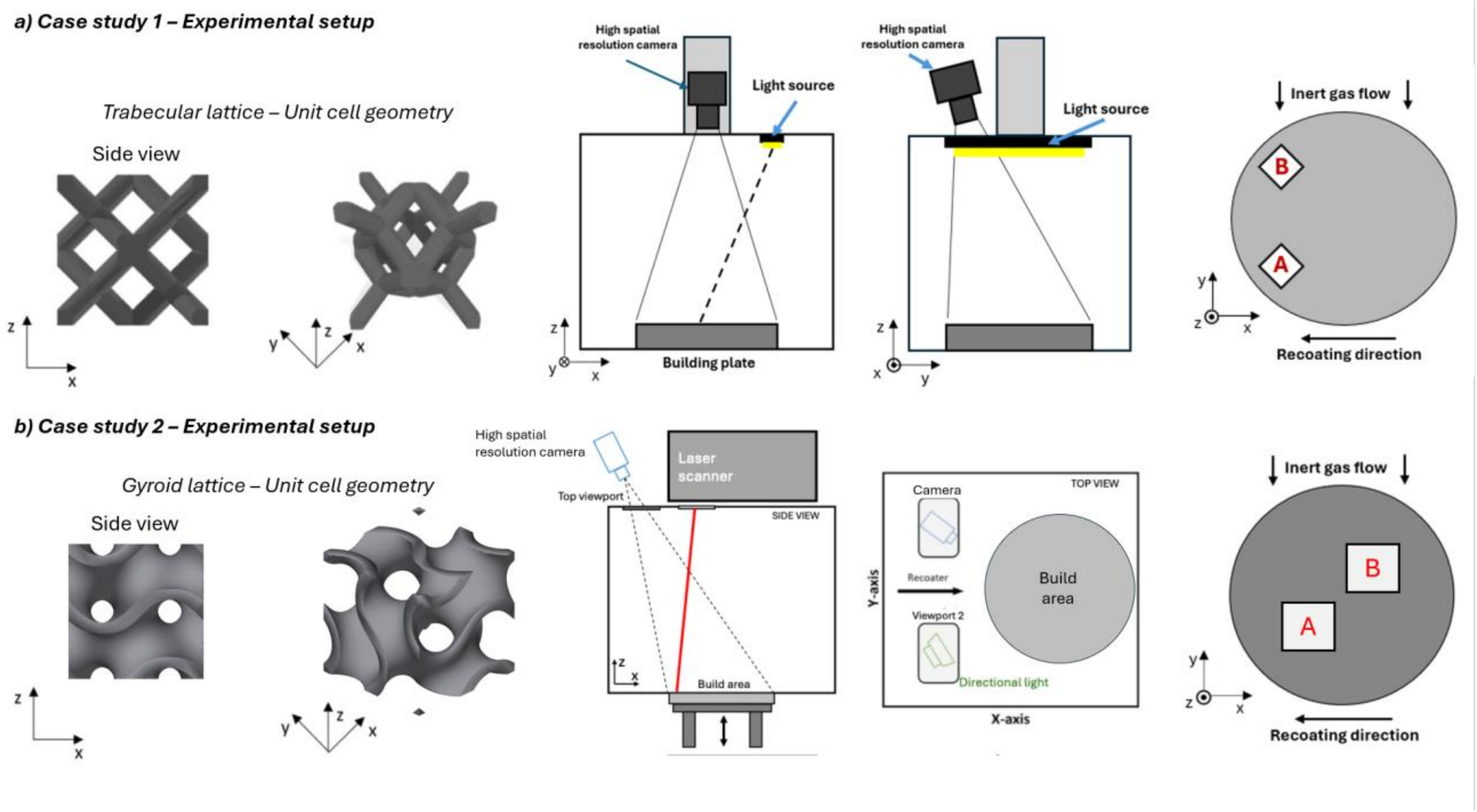

**Fig. 3.** Experimental settings in a) case study 1 and b) case study 2. On the left: side view and 3-D view of the 10 × 10 × 10 mm unit cell that composes the lattice structures in each case study. On the right: views of the camera and lighting setups and specimen location inside the build area.

In each case study, the analysis was performed on two copies of the same lattice specimen, hereafter referred to as specimen A and specimen B. Their respective locations within the build area of the two manufacturing setups are shown in Fig. 3, right panels. Due to their relative position with respect to the illumination system, these locations were selected as they were characterized by different powder-bed lighting conditions. Another source of variability arises from the layer-wise rotation of the laser scan direction, which, according to standard industrial practice, was set to 67° between consecutive layers, and thus affects how light is reflected toward the camera.

All as-built specimens were inspected using a North Star Imaging X25 X-ray CT scan system with a voxel size of 33 μm in case study 1, and 40 μm in case study 2. The segmentation of the CT data was performed using a standard histogram-based thresholding approach implemented in VGStudio MAX (Volume Graphics GmbH). The threshold value was selected based on the bimodal distribution of voxel intensities and applied as consistently as possible across all samples acquired under the same CT-scanning conditions. After thresholding, the volumes were exported in STL format and aligned to the nominal geometry using the Iterative Closest Point (ICP) algorithm implemented in MeshLab, following the same procedure described in [12]. The aligned CT data were finally sliced to generate binary ground truth masks for every manufactured layer. Both nominal and ground truth masks were aligned to the *in situ* images via landmark-based registration as part of camera calibration.

X-ray CT represents the only viable technique for non-destructive 3D characterization of complex shapes like lattice structures. In this study, CT data were used as a reference to assess the *in situ* geometry reconstruction accuracy. Based on these measurements, no defects or significant dimensional deviations were observed among the manufactured samples. Therefore, any differences affecting the *in situ* segmentation can be attributed mainly to variability in the *in situ* image properties and the resulting impact on segmentation accuracy.

**TABLE II** – Process parameters adopted in the two PBF-LB/M case studies

| Case study 1 | | | | | |
|---|---|---|---|---|---|
| Layer thickness | Power | Scan speed | Hatch distance | Scan mode | Scan strategy |
| 0.05 mm | 275 W | 1200 mm/s | 0.09 mm | Continuous mode | Meander |
| Case study 2 | | | | | |
| Layer thickness | Power | Scan speed | Hatch distance | Scan mode | Scan strategy |
| 0.03 mm | 320 W | 1200 mm/s | 0.15 mm | Continuous mode | Meander |

### *3.2 Numerical settings*

The training datasets were constructed using *in situ* images paired with ground truth masks derived from CT scans.

The experimental design was conceived with a dual objective. First, to systematically investigate how variability between training and test conditions affects segmentation accuracy. Second, to emulate an industrially relevant scenario, where the training effort and costs shall be minimized. Because of this, the training phase was purposely limited to image data from the initial portion of the available lattice structures, while subsequent layers of the same components were used as test dataset for the analysis and comparison of performances. Two distinct training scenarios were investigated, namely separate training and joint training, designed to expose different levels of photometric variability between training and inference data (Fig. 4).

In the separate training scenario, two independent models were trained, one for each specimen. Each model was trained using the first two complete stacks of unit cells along the build direction for each respective specimen (either A or B). This choice ensures that the training data span the vertical development of unit cells at two different build heights, thereby embedding both vertical and in-plane geometrical variability within training data. The corresponding inference set consisted of the remaining layers of both specimens in the two upper stacks of unit cells.

In the joint training scenario, a single model was trained using combined data from both lattice specimens A and B, specifically the first stack of unit cells at the bottom of each specimen, to keep the training dataset size consistent with the previous scenario. This configuration exposes the model to data acquired at different spatial locations on the build plate, thereby incorporating variability in illumination and imaging conditions already during training. The inference set consisted of the remaining layers from both specimens, allowing evaluation under broader and more heterogeneous conditions.

The way in which images from different portions of the lattice structures were organized into different training and inference datasets is depicted and summarized in Fig. 4.

To quantitatively assess the performance of the segmentation methods, the reconstructed binary masks were compared against the corresponding ground truth. The segmentation performance was quantified by comparing the reconstructed masks with the corresponding ground truth masks on a pixel-by-pixel basis. Each pixel was classified as either foreground (solidified material) or background (powder), and the agreement between prediction and ground truth was evaluated accordingly.

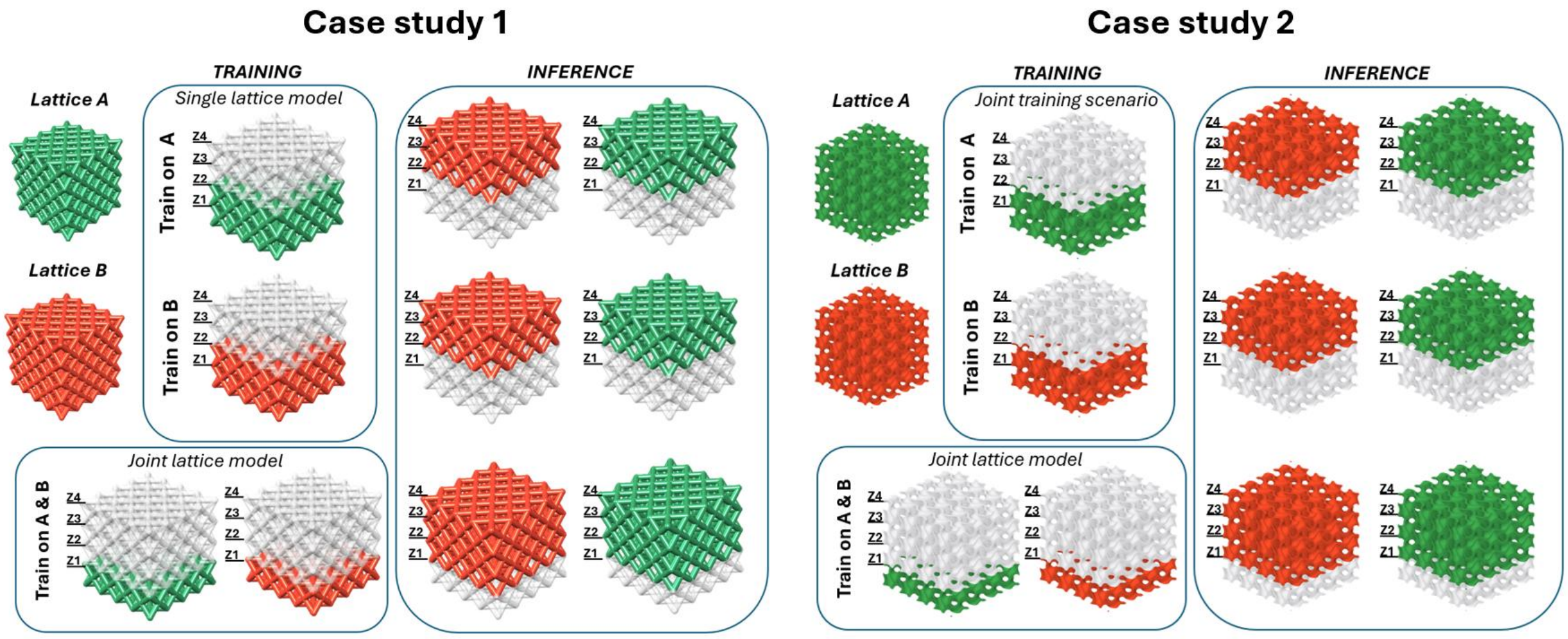

**Fig. 4.** Schematic representation of the dataset splits adopted for training and testing across the two explored case studies and in different training scenarios: separate training and joint training.

Pixel-wise accuracy [57] was adopted as the primary evaluation metric, but results in terms of intersection over union (IoU) [61] are reported as well. Pixel-wise accuracy measures the overall proportion of correctly classified pixels, accounting for both foreground and background regions, and thus provides a global assessment of segmentation performance. Conversely, IoU quantifies the overlap between the predicted and reference foreground regions, defined as the ratio between their intersection and union.

More specifically, each pixel contributes to one of four categories:

- *true positive (TP):* pixel-wise match between *in situ* segmentation (foreground) and ground truth,
- *true negative (TN),* pixel correctly labelled as background in the *in situ* segmentation,
- *false positive (FP),* pixel wrongly labelled as foreground in the *in situ* segmentation,
- *false negative (FN),* pixel wrongly labelled as background in the *in situ* segmentation.

The accuracy and the IoU, both ranging between 0 and 1, were computed as:

$$\text{Accuracy} = \frac{TP + TN}{TP + TN + FP + FN}, \qquad \text{IoU} = \frac{TP}{TP + FP + FN}$$

Reporting both metrics therefore provides complementary insights into overall pixel classification performance and foreground reconstruction quality.

The UNet-GNN approach was compared against two benchmarks discussed above, namely active contours and original U-Net. As regards the active contours methodology, we referred to the formulation previously applied to PBF-LB/M images in [14] which combined edge- and region-based terms. Active contours require no training, as the segmentation only relies on two inputs: i) the current powder bed image to be segmented, and ii) the starting contour to be used as level set 0, i.e., the nominal shape in the layer. However, active contours involve parameters to be set. Thus, images enclosed in the training set were used to estimate the optimal values of the two most relevant parameters, namely the weight $w$, which balances the relative contribution of edge- and region-based terms, and the parameter $r_{kernel}$, which is the radius of the kernel used for local spatial weighting around the evolving contour [14], [60]. The weight $w$ is such that $0 \leq w \leq 1$, where $w = 0$ corresponds to a purely edge-based segmentation, whereas $w = 1$ to a purely region-based one. The kernel is essentially associated to the region-based term, since the edge-based term exploits a global gradient function. According to [14], a dense stencil of width $1 + 2r_{kernel}$ was used. The search for optimal values of $w$ and $r_{kernel}$ was performed by means of a full grid search in the ranges $0 \leq w \leq 1$ and $1 \leq r_{kernel} \leq 12$ (expressed in number of pixels), with a step of 0.05 for $w$ and a step of 1 for $r_{kernel}$. More details about the active contour calibration procedure can be found in [60].

As regards the second competitor, i.e. the U-Net, the original encoder–decoder design proposed in [55] was adopted, with the same encoder–decoder structure as the UNet-GNN but without the graph-based module, i.e., with a purely convolutional bottleneck. To keep the comparison fair, the same hyperparameters were used for both the U-Net and the UNet-GNN (summarized in Table III). Moreover, the same training and inference settings were applied to both networks.

**TABLE III** - Training configuration shared by U-NET and UNet-GNN

| **Parameter** | **Value** |
|---|---|
| Input image size | 512 × 512 |
| Encoder base channels | 64 |
| Loss function | Combined BCE-Dice |
| Optimizer | AdamW |
| Learning rate | $1 \times 10^{-3}$ |
| LR scheduler | ReduceLROnPlateau |
| Batch size | 8 |
| Type of normalization | Batch normalization |
| Max. number of epochs | 100 |
| Early stopping patience | 10 epochs (min $\Delta = 1{\times}10^{-4}$) |
| Train/validation split | 80% / 20% |

In addition to testing and comparing methods in the presence of real data and real data variability sources, one additional analysis involved the introduction of different types of data perturbations to further assess the OOD generalization and accuracy of compared methods in the presence of controlled photometric variations. Such variations are representative of conditions that may be faced where the same *in situ* inspection method is deployed on different machines or in different productions sites, where different imaging and/or illumination configurations are present.

This analysis involved segmentation models trained on original image data in the absence of any applied perturbation, while three distinct types of perturbations were introduced in the inference stage:

- *Gamma perturbation ($\gamma$):* a nonlinear intensity transformation was introduced by modifying the gamma value of the images, altering contrast and brightness characteristics to simulate variations in illumination conditions. Values of $\gamma$ progressively deviating from 1 increase the perturbation severity.
- *Gaussian noise ($\sigma$):* random noise sampled from a Gaussian distribution $\mathcal{N}(0, \sigma^2)$ was added to the images, simulating decreasing levels of signal-to-noise ratios, representative of decreased image quality conditions. Increasing values of $\sigma$ correspond to higher noise severity.
- *Pixelation ($s$):* spatial resolution degradation was simulated by down-sampling the images by a scale factor $s \in (0,1]$ and subsequently up-sampling them back to the original image size, producing a loss of fine structural details, representative of conditions faced when using a lower resolution camera. Smaller values of $s$ correspond to more severe pixelation effects.

For each perturbation type, three levels of severity were applied. The specific parameter settings adopted for each severity level are summarized in Table IV. Fig. 5 shows examples of perturbations at different severities for one sample powder bed image showing one lattice cell in case study 1.

**TABLE IV** – Controlled perturbation levels

| Perturbation type | Perturbation level | Value |
|---|---|---|
| Gaussian noise | Low | σ = 5 |
| | Mid | σ = 10 |
| | High | σ = 20 |
| Gamma perturbation | Low | γ = 0.8 |
| | Mid | γ = 1.2 |
| | High | γ = 1.5 |
| Pixelation | Low | S = 0.75 |
| | Mid | S = 0.5 |
| | High | S = 0.25 |

Although the introduced perturbations are synthetic, they emulate several variability sources commonly encountered in industrial PBF-LB/M. Gamma transformations reproduce changes in global illumination and camera exposure, additive Gaussian noise approximates sensor noise and acquisition disturbances, whereas pixelation mimics the reduced spatial resolution that may arise when cameras with different optical configurations or larger fields of view are employed. Consequently, the proposed perturbation analysis represents an approximation of operating conditions under which *in situ* inspection may be deployed across different machines or production sites.

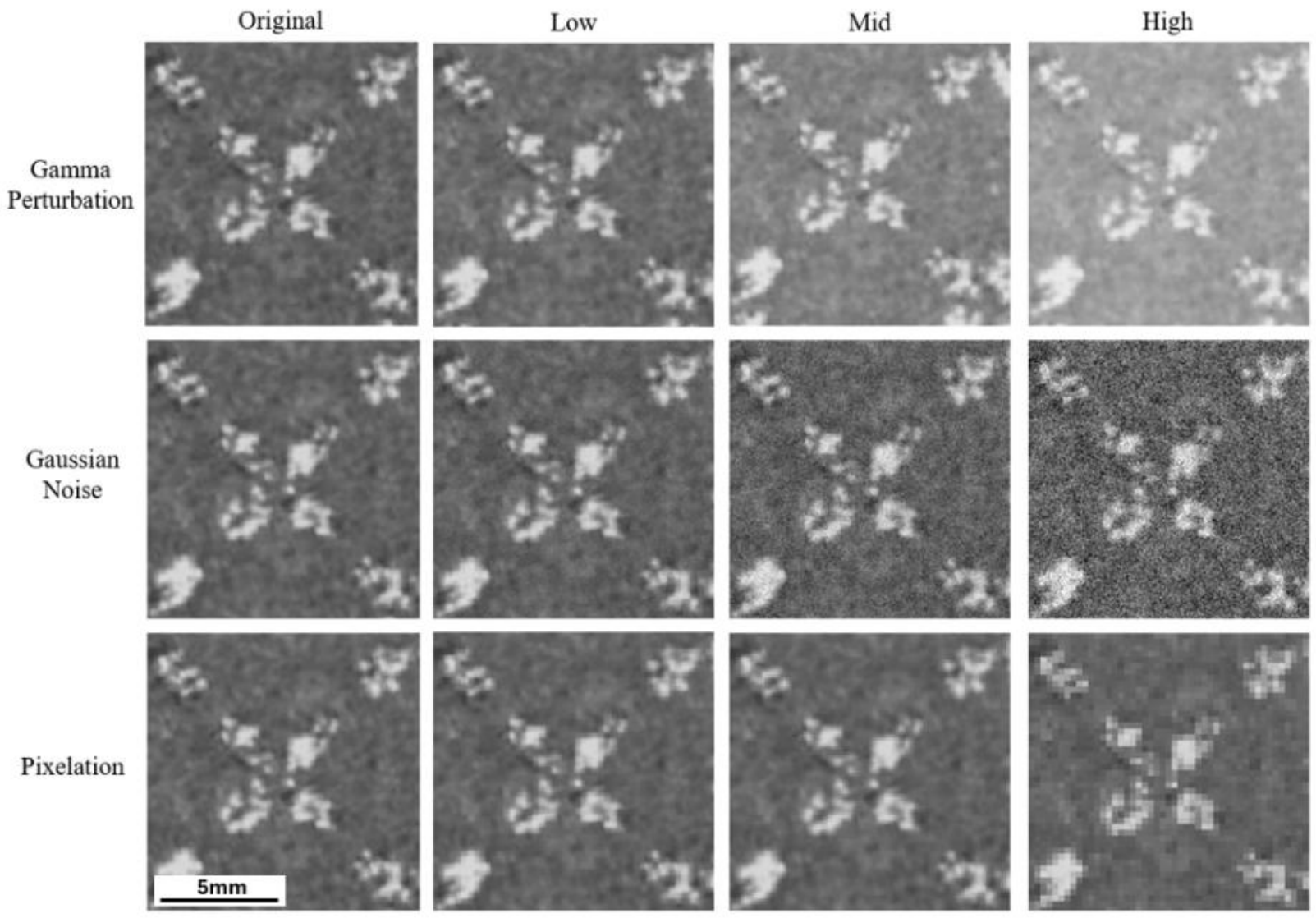

**Fig. 5.** Examples of data perturbations applied to the *in situ* image (one unit cell of sample A in case study 1). From top to bottom: gamma perturbation, Gaussian noise, and pixelation. For each perturbation type, three increasing levels of severity are shown from left to right, labelled as “low”, “mid”, and “high”.

*3.3 Artificial anomaly injection*

In addition to the image perturbations presented above, this study also included a preliminary exploration of detection performances in the presence of deviations from the nominal shape that, although being artificially injected, are representative of local deficiencies of material that can occur in PBF-LB/M. Indeed, this kind of anomaly may arise when portions of the nominal cross-section are not completely melted and consolidated. This may result from insufficient local energy input, laser interruptions, scan path deviations, powder-bed irregularities, or recoating-induced defects that prevent the intended material consolidation.

In this study, a set of additional layers in case study 1 was produced by forcing a local lack of material using a modified nominal geometry as shown in Fig. 6. More specifically, anomalies with two different severities were generated. The so-called “mid-severity” one consisted of halving the laser-exposed area of one single strut, whereas the “high-severity” anomaly consisted of a complete lack of exposed area in correspondence of one strut, resulting in an entirely missing geometrical feature.

Two layers per severity level were considered for each lattice specimen in case study 1, with anomalies injected in the same unit cell. The layers with artificial anomalies were then used to replace original layers manufactured in nominal conditions. These anomalous layers were considered to test the automated deviation detection performances.

Fig. 6 shows the nominal mask of a single unit cell, alongside the corresponding modified masks used to induce the anomaly at the two severity levels.

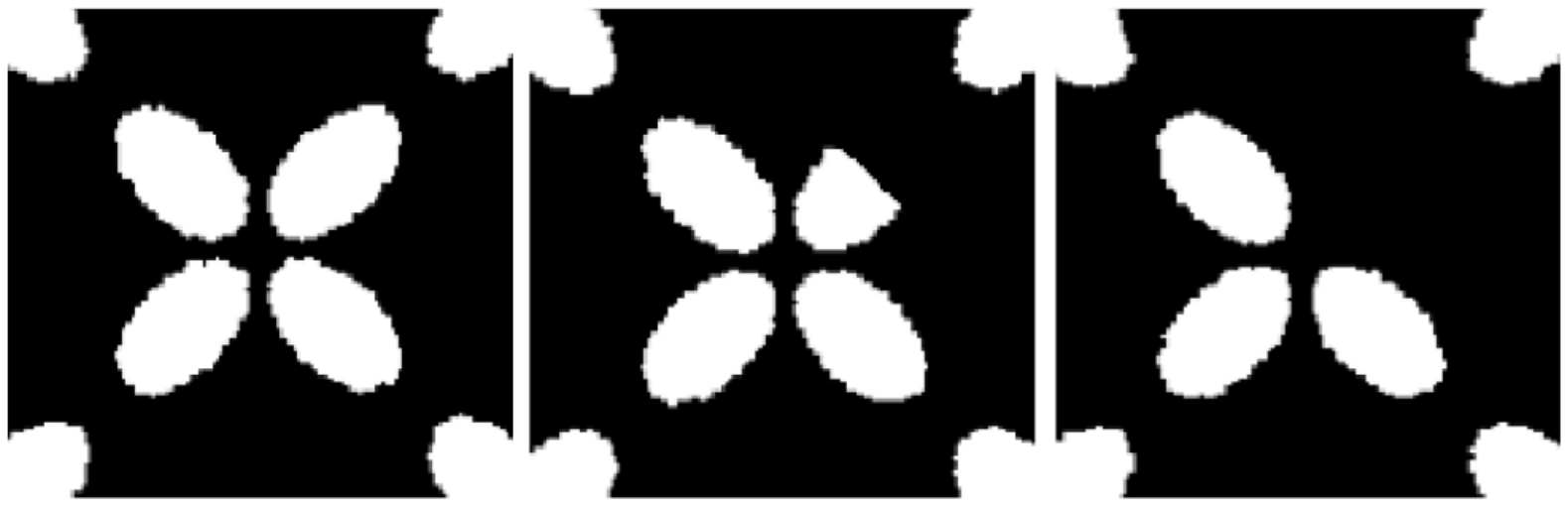

**Fig. 6.** Nominal mask of a single unit cell of trabecular lattice in case study 1 (left), and the corresponding modified masks used to induce the mid-severity (center) and high-severity (right) anomalies.

## IV. Results and discussion

The experimental results are presented and discussed in the following subsections.

Subsection 4.1 presents the *in situ* geometry reconstruction performance on original powder bed images along with the comparison against competitor methods. Subsection 4.2 presents an additional comparison in terms of OOD generalization to controlled data perturbations. Subsection 4.3 explores the capability of the tested methods to detect local geometrical anomalies unseen during training. Subsection 4.4 presents an ablation study together with additional considerations about the computational efficiency of tested methodologies.

*4.1 In situ reconstruction performance*

This section presents *in situ* image segmentation and geometry reconstruction results for the UNet-GNN model and its competitors across the two case studies. Fig. 7 shows the segmentation accuracy and IoU in successive layers along the build direction over the entire dataset in case study 1 in both training scenarios, namely separate training (i.e., models trained on one lattice and tested on either the same lattice or another lattice) and joint training (model jointly trained on both lattice A and lattice B and tested on remaining layers of both lattices). Fig. 8 shows the segmentation accuracy and IoU along the build direction over the entire dataset in case study 2 in the same two training scenarios. All panels in Fig. 7 and Fig. 8 refer to the reconstruction accuracy and IoU in the inference datasets only.

Fig. 7 and Fig. 8 show that when training and inference are performed on the same dataset (Train on A–Inference on A and Train on B–Inference on B), both deep learning approaches consistently outperform active contours, with UNet-GNN generally achieving the highest segmentation performance. The improvement over conventional U-Net is relatively limited in these conditions, particularly for dataset A, suggesting that both architectures can effectively reconstruct the nominal geometry when the inference images exhibit characteristics similar to those encountered during training. In contrast, active contours exhibit substantially larger layer-to-layer fluctuations and pronounced performance drops, particularly in correspondence of specific geometric configurations.

More pronounced differences between U-Net and UNet-GNN emerge when training and inference are performed on different datasets. As regards case study 1, when models trained on dataset B are applied to dataset A, the UNet-GNN maintains substantially higher accuracy and IoU values than the conventional U-Net across most of the investigated layers. A similar, although less pronounced, advantage is observed in the opposite configuration (Train on A–Inference on B). These findings suggest that incorporating graph-based feature representations can improve the ability of the segmentation model to generalize across imaging conditions not adequately represented in the training dataset, reducing its sensitivity to dataset-specific photometric characteristics.

Joint training on datasets A and B in case study 1 (Fig. 7) generally improves the consistency of reconstruction performance across the two datasets, particularly for the conventional U-Net. Under these conditions, the performance gap between U-Net and UNet-GNN becomes considerably smaller, with the two architectures exhibiting similar trends and segmentation performance across most layers. This suggests that increasing the diversity of the training data can partially compensate for the limitations of conventional convolutional feature extraction.

The results obtained for case study 2 (Fig. 8) largely confirm the findings discussed for case study 1. Both deep learning approaches consistently outperform active contours, while the UNet-GNN generally achieves the highest accuracy and IoU values across the investigated training and inference configurations. In particular, the advantage of graph augmentation is also observed under cross-dataset inference conditions, although the performance differences with respect to conventional U-Net are generally less pronounced than in case study 1. Joint training on both datasets tends to reduce these differences, further supporting the complementary roles of training-data diversity and graph-based feature representations.

The pixel-wise accuracy and IoU exhibit broadly consistent trends, supporting the effectiveness of the proposed UNet-GNN architecture in improving the OOD generalization of *in situ* geometric reconstruction to variations in image appearance, with the most significant benefits emerging under conditions of limited training-data representativeness.

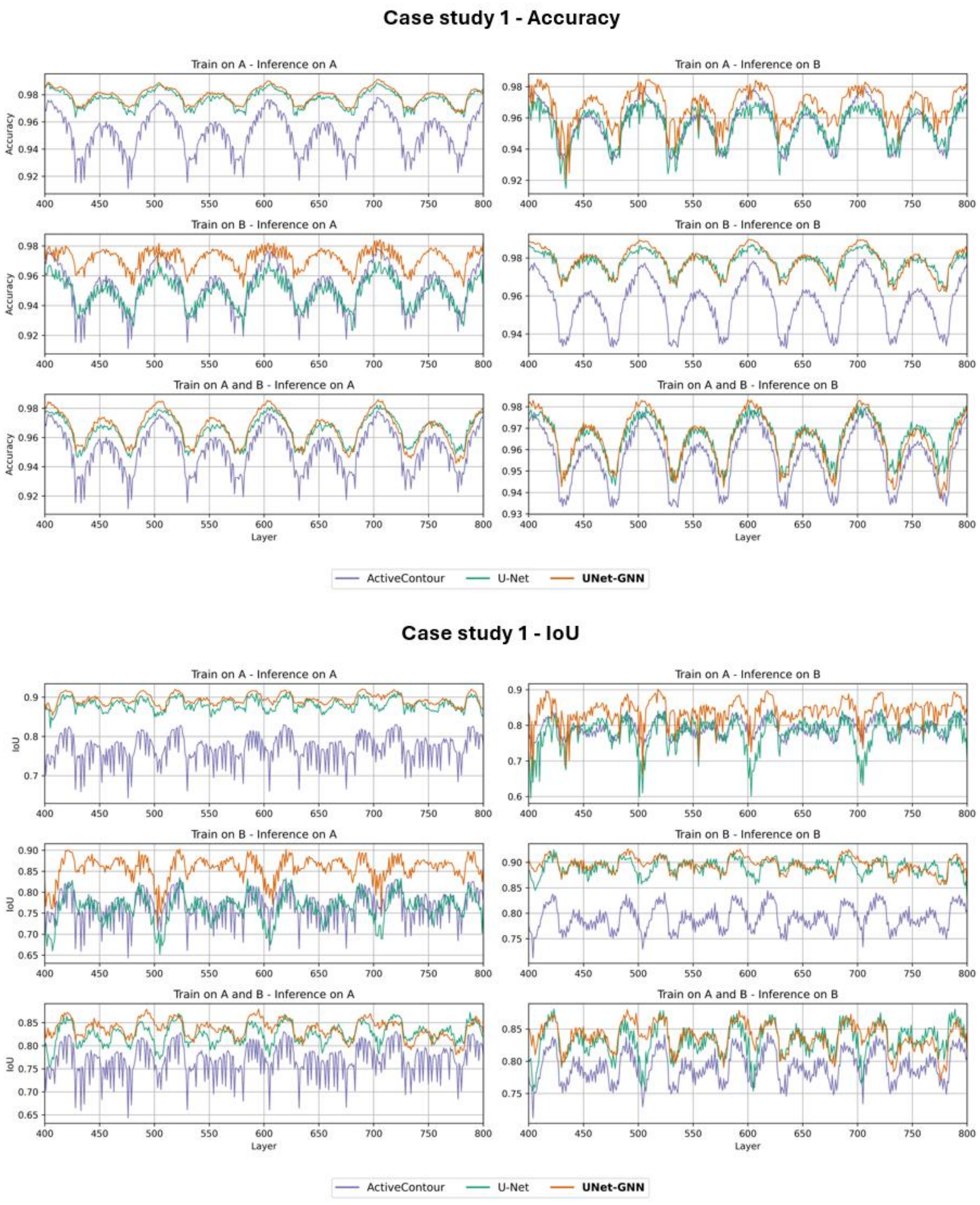

**Fig. 7.** Segmentation accuracy (top) and IoU (bottom) along the build direction in case study 1 under different training setups: i) models trained on lattice A and tested on either lattice A or lattice B, ii) models trained on lattice B and tested on either lattice A or lattice B, and iii) models jointly trained on both lattice A and lattice B and tested on remaining layers of lattice A and B. Results are reported for 1) active contour, 2) U-Net, and 3) UNet-GNN.

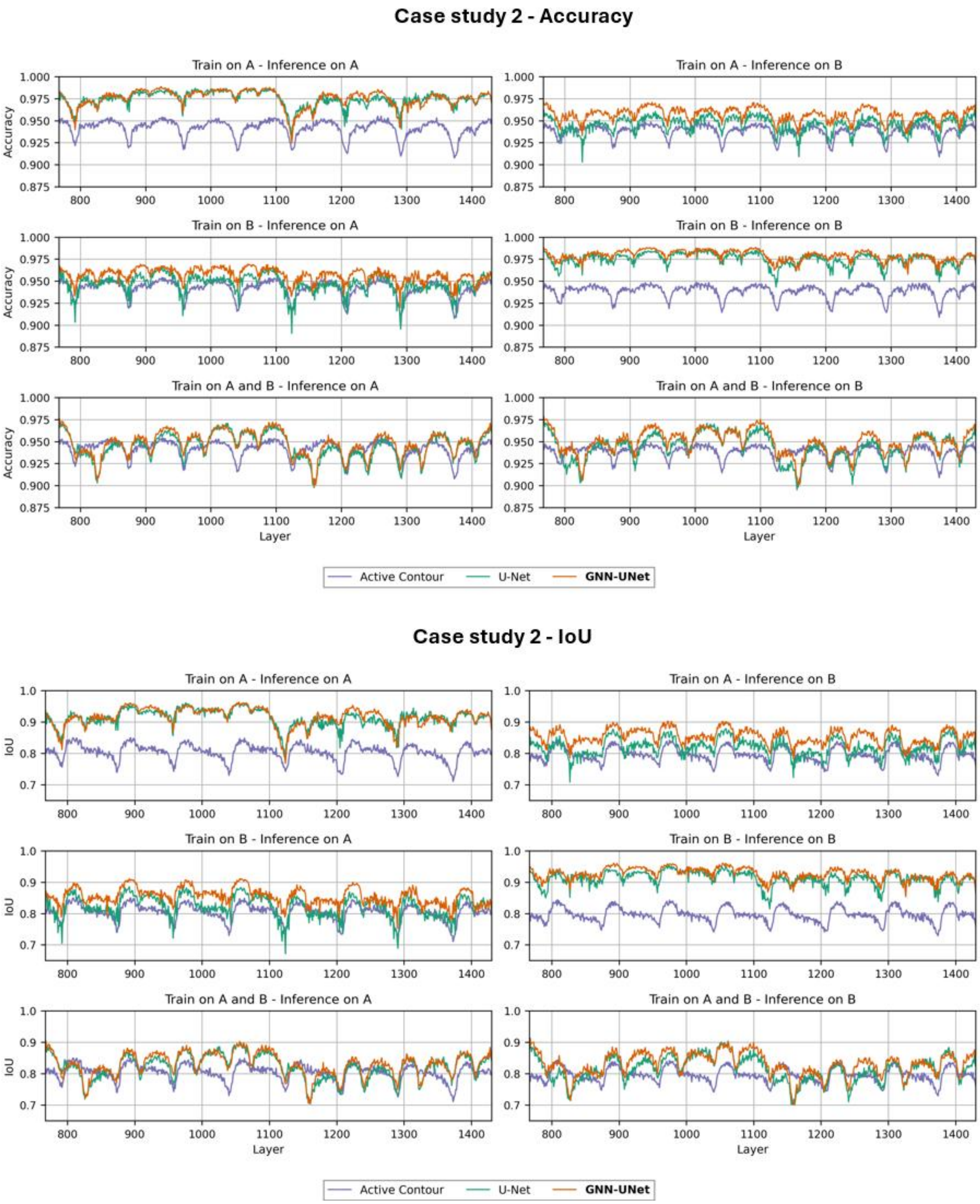

**Fig. 8.** Segmentation accuracy (top) and IoU (bottom) along the build direction in case study 2 under different training setups: i) models trained on lattice A and tested on either lattice A or lattice B, ii) models trained on lattice B and tested on either lattice A or lattice B, and iii) models jointly trained on both lattice A and lattice B and tested on remaining layers of lattice A and B. Results are reported for 1) active contour, 2) U-Net, and 3) UNet-GNN.

Fig. 7 and Fig. 8 also highlight that the accuracy systematically varies from layer to layer following a periodic pattern. This periodicity essentially depends on the geometrical evolution of unit cells' sections along the build direction. The influence of the trabecular lattice geometry (case study 1) on such repeating patterns of the segmentation accuracy is highlighted in more detail in Fig. 9.

Fig. 9 (right panel) illustrates sample images corresponding to half of a unit cell, spanning 100 successive layers, together with the associated segmentation accuracy profiles from the three compared methods (left panel). Due to the geometric symmetry of the unit cell along the build direction, the remaining half exhibits a mirrored behavior.

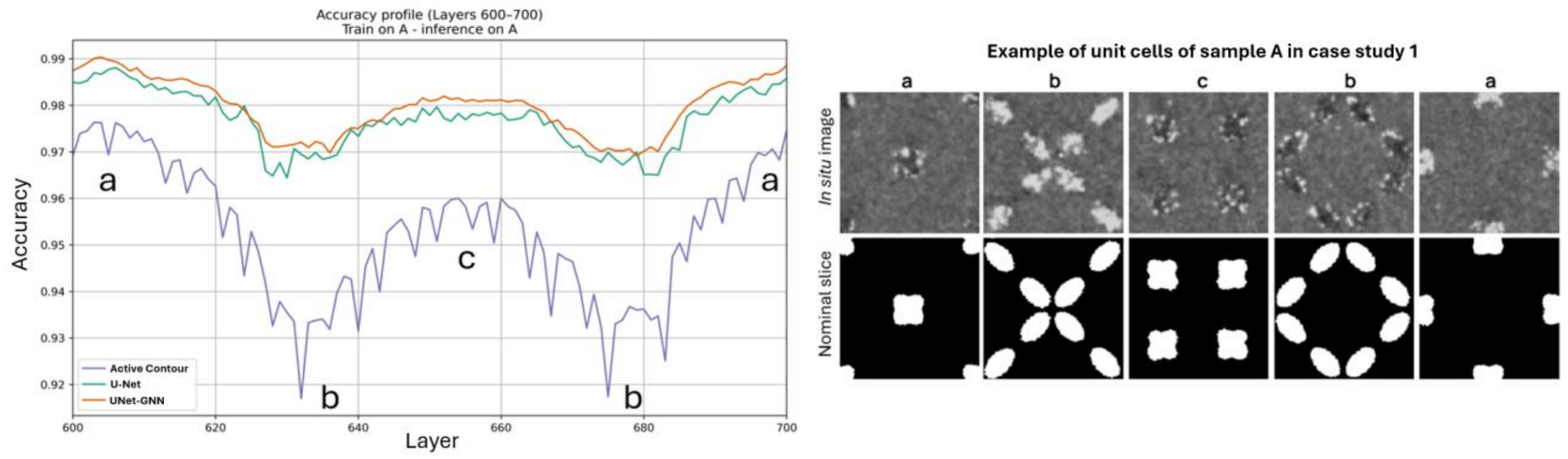

**Fig. 9.** Detail of segmentation accuracy profiles over half of a unit cell (layers 600–700) in case study 1 for active contour, U-Net, and UNet-GNN; the accuracy refers to models trained on lattice A and tested on the same lattice (left). Regions corresponding to distinct geometric features of the lattice cell: representative *in situ* images and nominal slices are reported for selected layers (right).

Three regions can be identified in Fig. 9: regions (a) correspond to layers where the outer nodes of the structure are manufactured: in these layers, all segmentation methods exhibit their respective highest accuracy; regions (b), on the contrary, correspond to sections of struts, where all methods exhibit their respective lowest accuracy; region (c), eventually, represents an intermediate condition, where struts converge into four nodes and intermediate accuracy values are observed.

In addition to these geometry-driven variations, the accuracy pattern also exhibits some higher frequency fluctuations. This is particularly evident looking at the accuracy of the active contour (Fig. 9, left panel). The seasonality of these valleys, where the segmentation accuracy drops, reflects the 67° offset of the layer-wise scan direction variation, and hence it is mainly driven by the way in which the solidified surface reflects the light towards the camera. Such scan direction-induced variations strongly affect the active contour performance, while the influence is lower as regards the U-Net performance and is further reduced in the presence of the UNet-GNN.

Upon the aforementioned subdivision into geometrical features, 95% confidence intervals for the mean accuracy of each method are reported in Fig. 10 in two distinct portions of the build in case study 1, corresponding either to nodes (a) or struts (b). For "nodes", a single confidence interval is computed using the accuracy values from all 20-layer windows centered at layers 400, 500, 600, 700, and 800, where the windows at the boundary layers are truncated. For "struts", the confidence interval is analogously computed using the accuracy values from the 20-layer windows centered at layers 625, 675, 725, and 775, thus considering the same total number of layers as for nodes. For enhanced visualization purposes, the y-axis range in Fig. 10 is independently scaled for each subplot to emphasize inter-method differences.

For conciseness, Fig. 10 presents the comparison of segmentation performance across different geometric features for case study 1, focusing on the segmentation accuracy as performance metric. The corresponding IoU results, along with both accuracy and IoU results for case study 2, are provided in the Appendix, as they support the same overall findings.

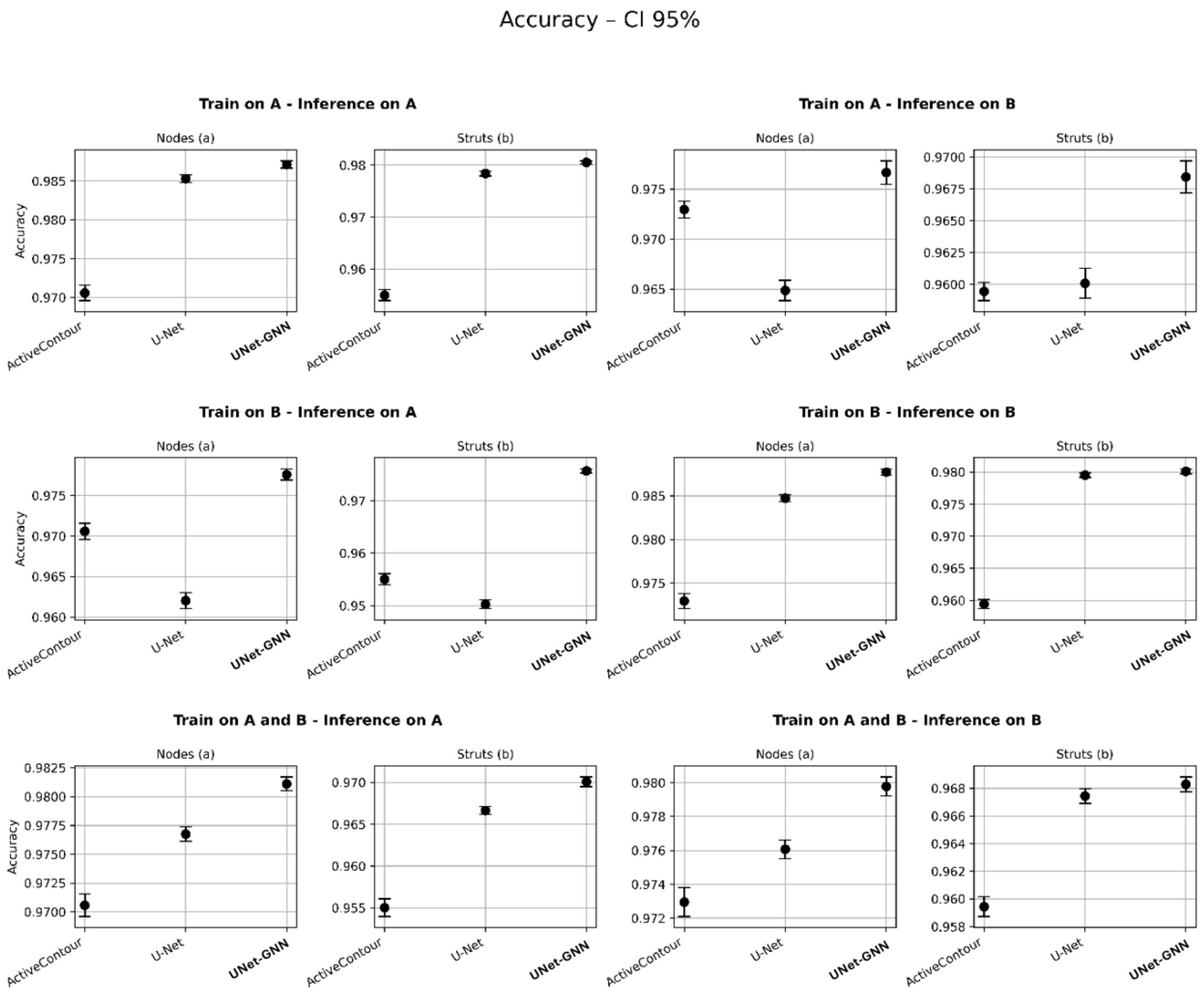

**Fig. 10.** 95% confidence intervals for the mean accuracy computed over regions a (nodes) and b (struts) of lattice structures in case study 1 for all compared segmentation methods. The y-axis range is independently scaled for each subplot to emphasize inter-method differences.

The results in Fig. 10 show that the proposed UNet-GNN generally achieves the highest pixel-wise accuracy for both feature types across the different training and inference configurations, indicating that its benefits are not restricted to a specific geometric feature. The improvements are particularly evident under cross-dataset inference conditions, where conventional U-Net exhibits a more pronounced performance degradation in some configurations, achieving accuracy levels comparable to, or even lower than, those obtained with the active contour method.

Fig. 11 completes the analysis showing representative examples of the *in situ* contour reconstruction of a unit cell obtained using the two deep learning segmentation methods, namely the UNet-GNN and the U-Net. The corresponding ground truth is shown too, as a reference.

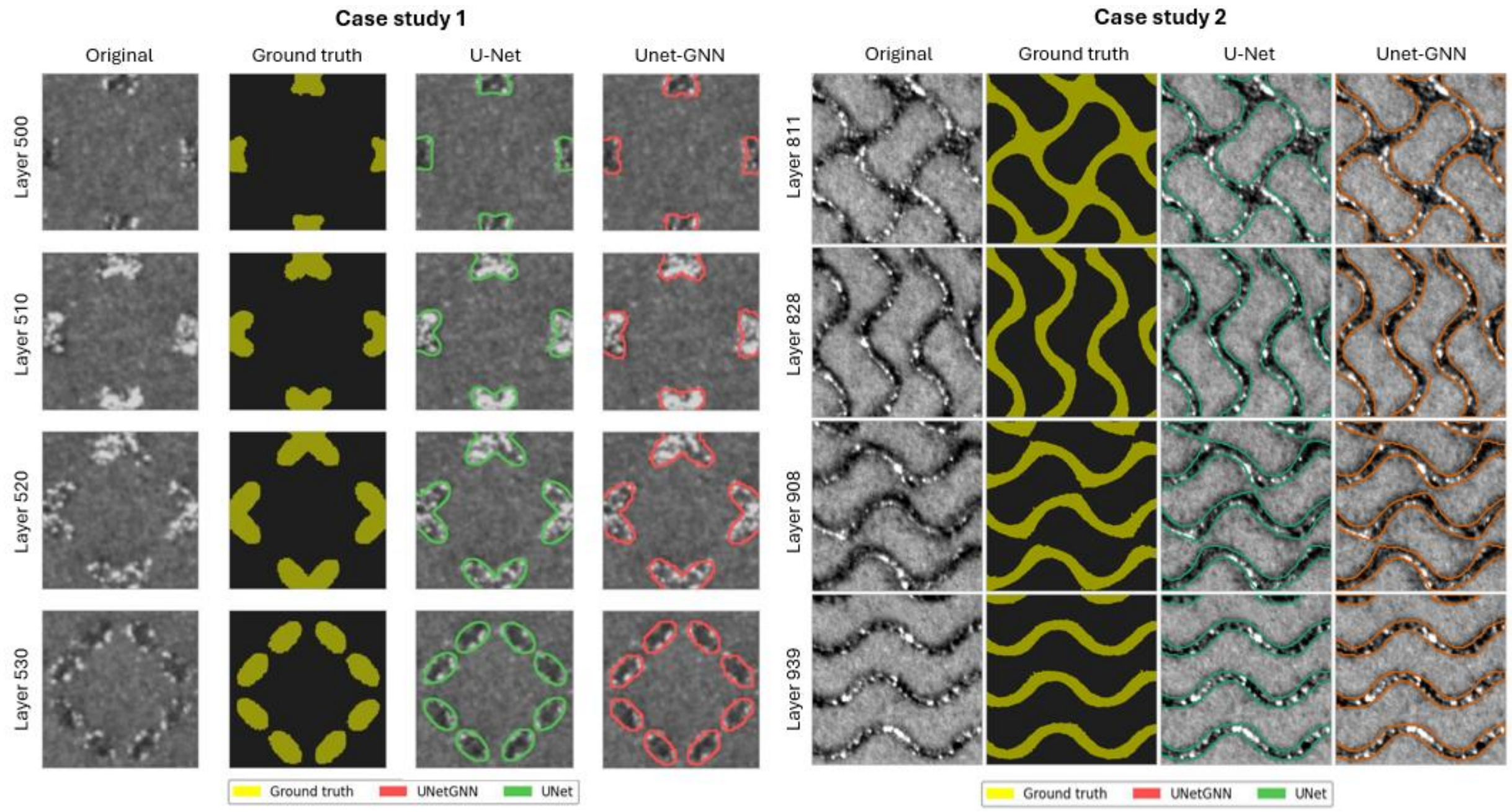

**Fig. 11.** Examples of original *in situ* images at different layers of specimen A (one single unit cell is shown) in case study 1 (left) and case study 2 (right), the corresponding ground-truth, and the segmentation obtained by using the U-Net and the UNet-GNN

Overall, the results can be summarized as follows:

1) The proposed UNet-GNN generally outperforms both conventional U-Net and active contours across the investigated training scenarios. Its advantage becomes particularly pronounced when training and inference are performed on different lattice specimens (i.e., training on lattice A and testing on lattice B, or vice versa), highlighting the potential of graph augmentation to improve robustness to photometric distribution shifts.
2) While both deep learning architectures achieve substantially higher segmentation accuracy than active contours when training and inference conditions are homogeneous, conventional U-Net exhibits a more pronounced performance degradation under cross-dataset inference, suggesting a greater dependence on the specific imaging characteristics represented in the training data.
3) When training data from both lattice specimens are combined, the performance gap between conventional U-Net and UNet-GNN generally decreases, indicating that greater training-data diversity can improve the robustness of conventional convolutional segmentation. Nevertheless, UNet-GNN maintains a generally favorable performance, suggesting that graph-based feature representations provide benefits complementary to those achieved by increasing training-data diversity.

4) The main findings are consistently confirmed across the two case studies. This consistency provides further evidence of the effectiveness of the proposed UNet-GNN architecture, suggesting that the benefits of graph augmentation are not restricted to a specific geometry, material, or manufacturing configuration.

These results indicate that a graph-augmented model may not only yield more accurate *in situ* geometry reconstruction than benchmark methods, including the basic U-Net configuration, but it also enhances the OOD generalization capability as photometric properties vary from training to inference, within the build area, and from layer to layer or from build to build.

From an industrial deployment perspective, such enhanced performance is particularly relevant as it may support more consistent performance of the *in situ* inspection tool across multiple builds and parts manufactured in different locations of the build area, while reducing the need to augment the size of the training set.

### *4.2 OOD generalization under controlled data perturbations*

To further investigate the OOD generalization of the method to variations in input image quality, three types of data perturbations were introduced on the test datasets. Under these conditions, the performance of the UNet-GNN was compared against the best-performing baseline model, namely the original U-Net. The active contours method, showing consistently lower accuracy across most tested conditions, was excluded from this additional comparison.

Fig. 12 shows the boxplots of the *in situ* reconstruction accuracy provided by the two methods under three different image perturbations. In both cases, models were trained on unperturbed images in case study 1 from lattice A (Fig. 12, left panel) and lattice B (Fig. 12, right panel), respectively, and then evaluated on perturbed test sets. The boxplots aggregate the accuracy values across layers corresponding to nodes and struts regions. Table V and Table VI, instead, summarize the performances dividing them with respect to the type of geometrical feature, i.e., node or strut, for all training and testing combinations and all types of image perturbations.

Overall, these results show that:

1) The UNet-GNN generally outperforms the U-Net in tested scenarios, although the gap between the two methods depends on the type of data perturbation (more evident for gamma perturbation and Gaussian noise addition);
2) in the presence of a gamma perturbation, which affects the global contrast and brightness properties of the image, the performance of both models degrades, but the UNet-GNN remains systematically more accurate than the competitor;
3) when Gaussian noise is added to the image, the U-Net exhibits an abrupt drop of the segmentation accuracy, while the UNet-GNN yields higher and more stable performance;
4) the reduction of the image resolution (pixelation scenario) has a lower impact on the segmentation accuracy across the tested severity levels, but also in this case the UNet-GNN performs steadily better than the U-Net.

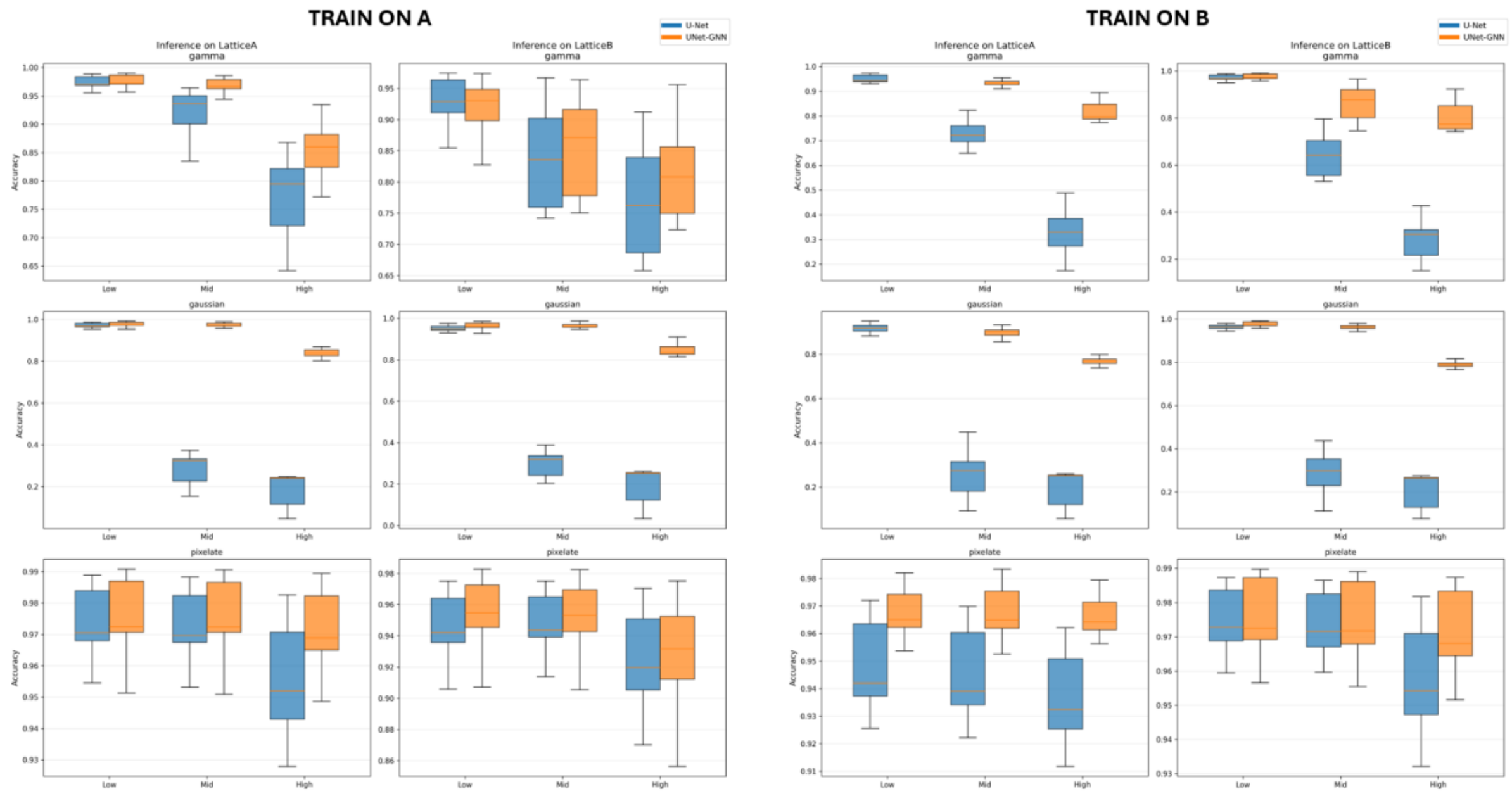

**Fig. 12.** Left: accuracy boxplots of U-Net and UNet-GNN models trained on lattice A (case study 1) and evaluated on either lattice A or lattice B of the same case study under different data perturbations (Gamma correction, Gaussian noise, and Pixelation). Right: accuracy boxplots of U-Net and UNet-GNN models trained on lattice B (case study 1) and evaluated on either lattice A or lattice B of the same case study under the same data perturbations. For each perturbation type, three severity levels (Low, Mid, High) are considered. The boxplots aggregate the accuracy values across layers corresponding to nodes and struts regions.

Fig. 12 shows that the relative improvement provided by the UNet-GNN depends on the nature of the perturbation. The largest performance gain is observed under gamma transformations and additive noise, both of which primarily alter local intensity statistics while leaving the underlying geometry unchanged. This observation is consistent with the hypothesis that graph-based feature processing reduces the model's reliance on local photometric patterns. Conversely, image pixelation removes geometric information from the input itself. Since this degradation directly affects the spatial information available to both networks, the improvement achievable through graph reasoning is smaller, although still consistently observable.

**TABLE V –** Mean accuracy and corresponding 95% confidence intervals for the UNet-GNN and the U-Net evaluated under different image data perturbation in case study 1 with different severity levels – separate training scenarios

| **Perturbation** | | **Nodes (a)** | | **Struts (b)** | |
|---|---|---|---|---|---|
| **Type** | **Level** | **U-Net** | **UNet-GNN** | **U-Net** | **UNet-GNN** |
| **Train on A, test on Lattice A** | | | | | |
| **Gamma** | Low | 0.985 [0.984–0.986] | **0.986 [0.984–0.987]** | 0.970 [0.969–0.970] | **0.972 [0.972–0.972]** |
| | Mid | 0.946 [0.943–0.948] | **0.980 [0.979–0.981]** | 0.914 [0.908–0.920] | **0.963 [0.962–0.964]** |
| | High | 0.821 [0.813–0.829] | **0.890 [0.886–0.894]** | 0.750 [0.738–0.761] | **0.836 [0.830–0.841]** |
| **Gaussian noise** | Low | 0.982 [0.981–0.984] | **0.986 [0.985–0.987]** | 0.966 [0.965–0.967] | **0.972 [0.972–0.973]** |
| | Mid | 0.215 [0.209–0.221] | **0.983 [0.982–0.984]** | 0.334 [0.332–0.336] | **0.970 [0.969–0.970]** |
| | High | 0.107 [0.103–0.112] | **0.857 [0.855–0.859]** | 0.241 [0.240–0.242] | **0.832 [0.830–0.835]** |
| **Pixelate** | Low | 0.984 [0.983–0.985] | **0.986 [0.985–0.987]** | 0.968 [0.968–0.969] | **0.972 [0.971–0.972]** |
| | Mid | 0.984 [0.983–0.985] | **0.986 [0.985–0.988]** | 0.960 [0.959–0.961] | **0.971 [0.970–0.972]** |
| | High | 0.971 [0.969–0.972] | **0.982 [0.981–0.983]** | 0.945 [0.944–0.947] | **0.965 [0.964–0.966]** |
| **Train on A, test on Lattice B** | | | | | |
| **Gamma** | Low | **0.965 [0.964–0.966]** | 0.955 [0.951–0.958] | **0.914 [0.910–0.917]** | 0.904 [0.898–0.911] |
| | Mid | 0.908 [0.902–0.914] | **0.913 [0.906–0.920]** | 0.795 [0.786–0.805] | **0.822 [0.810–0.833]** |
| | High | 0.850 [0.844–0.857] | **0.870 [0.864–0.876]** | 0.715 [0.707–0.723] | **0.772 [0.766–0.778]** |
| **Gaussian noise** | Low | 0.965 [0.963–0.966] | **0.977 [0.975–0.978]** | 0.946 [0.945–0.948] | **0.957 [0.956–0.958]** |
| | Mid | 0.234 [0.230–0.237] | **0.973 [0.970–0.975]** | 0.336 [0.332–0.340] | **0.949 [0.948–0.950]** |
| | High | 0.115 [0.110–0.119] | **0.868 [0.865–0.870]** | 0.253 [0.252–0.254] | **0.830 [0.829–0.832]** |
| **Pixelate** | Low | 0.964 [0.963–0.965] | **0.976 [0.975–0.978]** | 0.939 [0.937–0.940] | **0.950 [0.948–0.952]** |
| | Mid | 0.964 [0.963–0.965] | **0.974 [0.972–0.976]** | 0.938 [0.936–0.939] | **0.945 [0.942–0.948]** |
| | High | 0.950 [0.948–0.952] | **0.952 [0.950–0.954]** | 0.906 [0.904–0.908] | **0.913 [0.910–0.916]** |
| **Train on B, test on Lattice A** | | | | | |
| **Gamma** | Low | 0.960 [0.958–0.961] | **0.968 [0.966-0.970]** | 0.942 [0.941–0.943] | **0.951 [0.949-0.953]** |
| | Mid | 0.720 [0.713–0.726] | **0.939 [0.937–0.941]** | 0.735 [0.727–0.743] | **0.929 [0.928–0.931]** |
| | High | 0.243 [0.231–0.255] | **0.853 [0.851–0.856]** | 0.379 [0.369–0.388] | **0.791 [0.790–0.792]** |
| **Gaussian noise** | Low | 0.931 [0.929–0.933] | **0.933 [0.931-0.935]** | 0.912 [0.910–0.915] | **0.921 [0.919-0.924]** |
| | Mid | 0.163 [0.151–0.174] | **0.908 [0.906–0.911]** | 0.312 [0.304–0.320] | **0.892 [0.889–0.896]** |
| | High | 0.113 [0.109–0.117] | **0.758 [0.755–0.761]** | 0.251 [0.250–0.252] | **0.774 [0.772–0.776]** |
| **Pixelate** | Low | 0.962 [0.961–0.963] | **0.967 [0.966-0.968]** | 0.943 [0.942–0.944] | **0.967 [0.965-0.968]** |
| | Mid | 0.962 [0.960–0.963] | **0.965 [0.964–0.967]** | 0.940 [0.939–0.941] | **0.964 [0.964–0.965]** |
| | High | 0.960 [0.957–0.961] | **0.964 [0.962–0.966]** | 0.937 [0.936–0.938] | **0.964 [0.963–0.965]** |
| **Train B, test on Lattice B** | | | | | |
| **Gamma** | Low | 0.982 [0.980–0.984] | **0.986 [0.984–0.988]** | 0.966 [0.966–0.967] | **0.970 [0.970–0.971]** |
| | Mid | 0.694 [0.682–0.706] | **0.932 [0.926–0.938]** | 0.610 [0.596–0.625] | **0.832 [0.822–0.841]** |
| | High | 0.210 [0.202–0.218] | **0.858 [0.855–0.862]** | 0.333 [0.327–0.340] | **0.762 [0.760–0.764]** |
| **Gaussian noise** | Low | 0.972 [0.970–0.973] | **0.986 [0.984–0.988]** | 0.960 [0.958–0.961] | **0.970 [0.970–0.971]** |
| | Mid | 0.205 [0.189–0.222] | **0.969 [0.967–0.971]** | 0.335 [0.327–0.342] | **0.959 [0.958–0.960]** |
| | High | 0.123 [0.119–0.127] | **0.792 [0.789–0.795]** | 0.264 [0.263–0.265] | **0.786 [0.784–0.788]** |
| **Pixelate** | Low | 0.982 [0.981–0.984] | **0.986 [0.984–0.987]** | 0.970 [0.969–0.972] | **0.971 [0.970–0.972]** |
| | Mid | 0.982 [0.981–0.984] | **0.986 [0.984–0.988]** | 0.970 [0.968–0.972] | 0.971 [0.971–0.972] |
| | High | 0.971 [0.970–0.972] | **0.985 [0.983–0.987]** | 0.948 [0.947–0.949] | **0.970 [0.970–0.971]** |

**TABLE VI –** Mean accuracy and corresponding 95% confidence intervals for the UNet-GNN and the U-Net evaluated under different image data perturbation in case study 1 with different severity levels – joint training scenarios

| Perturbation | | Nodes (a) | | Struts (b) | |
|---|---|---|---|---|---|
| **Type** | **Level** | **U-Net** | **UNet-GNN** | **U-Net** | **UNet-GNN** |
| **Joint training (A and B), test on Lattice A** | | | | | |
| **Gamma** | Low | 0.944 [0.942–0.946] | **0.981 [0.980–0.982]** | 0.868 [0.864–0.873] | **0.963 [0.962–0.963]** |
| | Mid | 0.652 [0.644–0.660] | **0.678 [0.671–0.684]** | 0.672 [0.659–0.684] | **0.731 [0.727–0.734]** |
| | High | 0.193 [0.190–0.197] | **0.574 [0.571–0.577]** | 0.292 [0.290–0.293] | **0.668 [0.663–0.672]** |
| **Gaussian noise** | Low | 0.948 [0.947–0.950] | **0.981 [0.979–0.982]** | 0.874 [0.870–0.878] | **0.968 [0.967–0.968]** |
| | Mid | 0.942 [0.940–0.944] | **0.981 [0.979–0.982]** | 0.868 [0.863–0.872] | **0.968 [0.967–0.968]** |
| | High | 0.555 [0.549–0.560] | **0.884 [0.882–0.886]** | 0.587 [0.582–0.592] | **0.859 [0.857–0.861]** |
| **Pixelate** | Low | 0.947 [0.945–0.949] | **0.981 [0.980–0.983]** | 0.874 [0.870–0.878] | **0.967 [0.967–0.968]** |
| | Mid | 0.945 [0.943–0.947] | **0.981 [0.979–0.982]** | 0.873 [0.869–0.877] | **0.967 [0.966–0.968]** |
| | High | 0.943 [0.941–0.945] | **0.980 [0.977–0.981]** | 0.872 [0.868–0.876] | **0.960 [0.958–0.962]** |
| **Joint training (A and B), test on Lattice B** | | | | | |
| **Gamma** | Low | 0.946 [0.944–0.948] | **0.979 [0.978–0.981]** | 0.862 [0.857–0.867] | **0.957 [0.956–0.958]** |
| | Mid | 0.620 [0.610–0.629] | **0.643 [0.638–0.649]** | 0.570 [0.562–0.579] | **0.693 [0.689–0.697]** |
| | High | 0.178 [0.173–0.183] | **0.565 [0.562–0.567]** | 0.281 [0.277–0.284] | **0.656 [0.652–0.661]** |
| **Gaussian noise** | Low | 0.948 [0.946–0.950] | **0.979 [0.977–0.981]** | 0.868 [0.863–0.872] | **0.967 [0.966–0.968]** |
| | Mid | 0.947 [0.945–0.949] | **0.977 [0.976–0.978]** | 0.862 [0.857–0.866] | **0.959 [0.959–0.960]** |
| | High | 0.573 [0.569–0.577] | **0.932 [0.931–0.934]** | 0.582 [0.579–0.585] | **0.888 [0.886–0.890]** |
| **Pixelate** | Low | 0.949 [0.947–0.951] | **0.979 [0.977–0.981]** | 0.868 [0.864–0.873] | **0.967 [0.966–0.968]** |
| | Mid | 0.948 [0.946–0.950] | **0.979 [0.977–0.981]** | 0.869 [0.865–0.874] | **0.967 [0.966–0.968]** |
| | High | 0.946 [0.944–0.950] | **0.977 [0.976–0.980]** | 0.861 [0.860–0.863] | **0.960 [0.958–0.962]** |

This analysis highlights the higher potential of a graph-augmented methodology to preserve accurate and stable *in situ* segmentation performance under degraded image quality. This improved OOD generalization can be attributed to the ability of the graph-based representation to capture relational dependencies in the feature space and contextual information beyond local pixel intensity patterns, effectively aggregating features into a structured representation of the underlying geometry. By operating on relational information rather than purely pixel-wise features, the approach reduces sensitivity to local image properties, such as those induced by the imaging and illumination setup and those induced by the process itself. This, in turn, enables more consistent performance across varying acquisition conditions, alleviating the need for frequent retraining or recalibration of the model when undesired variations occur.

Notably, the reported performance was achieved with a relatively small training dataset. This is particularly relevant for industrial applications, as it reduces data acquisition and labeling efforts, thereby lowering implementation costs while maintaining desired accuracy.

### *4.3 Automated anomaly detection*

A preliminary evaluation of the capability of the proposed UNet-GNN to detect process-induced geometric deviations is presented in this section. Consistent with the previous analysis, the conventional U-Net was adopted as the benchmark method. For both segmentation approaches, the adaptive control chart described in Section 2.3 was combined with the models trained using the joint training strategy, i.e., using images from both specimen A and specimen B in case study 1. Similar results were obtained using specimen-specific models and are therefore omitted for brevity. During the test phase, the control chart was applied independently to the reconstructed geometries of specimens A and B. The resulting control charts are reported in Fig. 13.

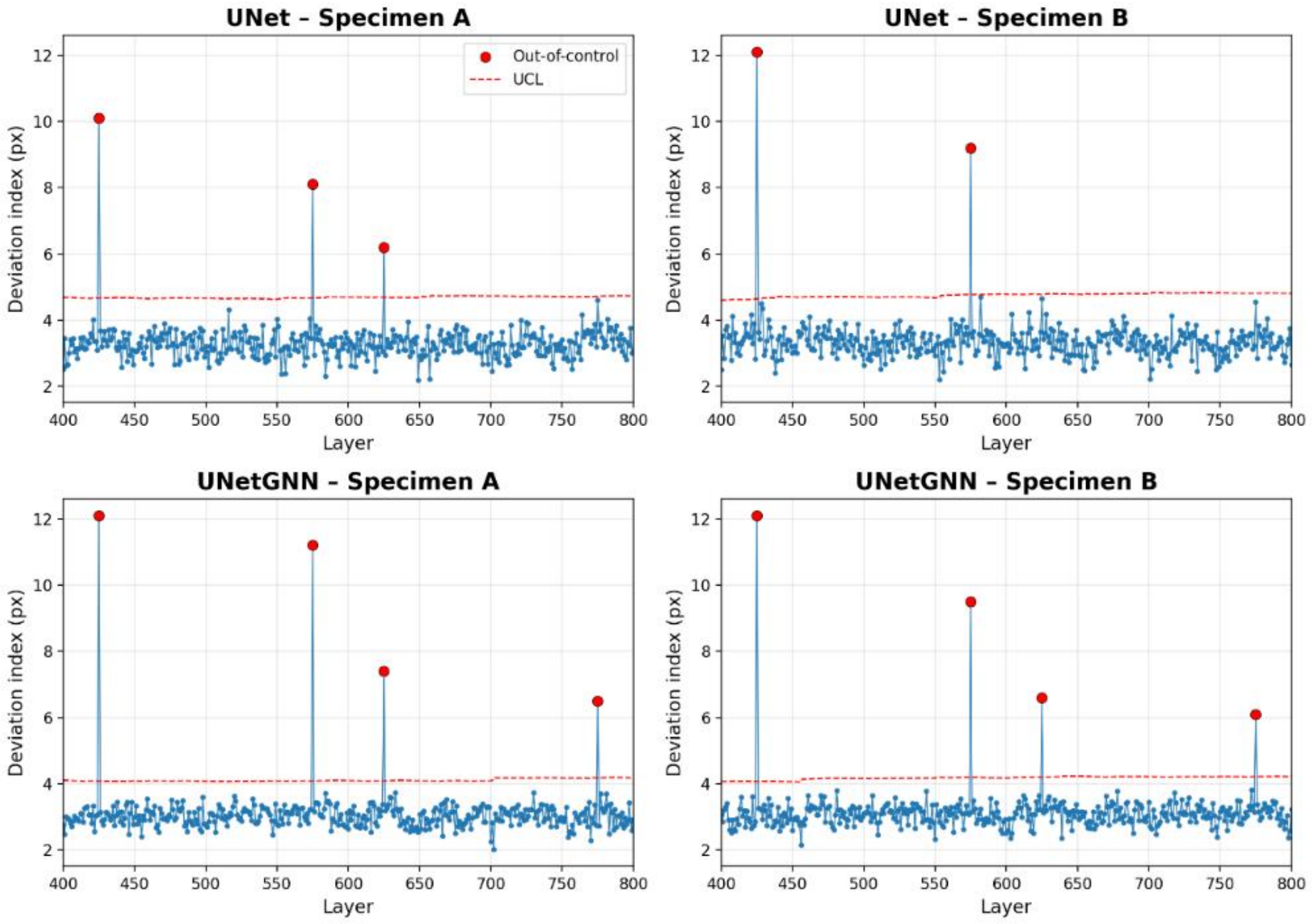

**Fig. 13.** Adaptive control charts based on the deviation index $d_i^{minmax}$ combined with the U-Net (top panels) and UNet-GNN (bottom panels) in the presence of seeded anomalies in case study 1. Left panels show the test phase of the control charts applied to layers belonging to specimen A, while right panels show the test phase of the control charts for specimen B (both models were trained adopting the joint training scenario). Mid-severity anomalies were included at layers 625 and 775, respectively, whereas high-severity anomalies were included at layers 425 and 575.

The top row of Fig. 13 shows the monitoring results obtained using the conventional U-Net, whereas the bottom row reports the corresponding results obtained using the proposed UNet-GNN. The target false alarm probability was set to $\alpha = 0.01$. No false alarms were generated by any of the four monitoring charts.

Artificial high-severity anomalies, corresponding to the complete removal of a lattice strut, were introduced at layers 425 and 575 in both specimens. These anomalies were successfully detected by both segmentation methods. Mid-severity anomalies, corresponding to the removal of approximately half of a strut, were introduced at layers 625 and 775. When the reconstructed geometry was obtained using the proposed UNet-GNN, both anomalies were consistently detected in the two specimens. In contrast, the control chart based on the conventional U-Net failed to detect one anomaly in specimen A and both anomalies in specimen B, despite the relatively large magnitude of the injected deviations (in the order of 600 – 700 µm, approximately twice the average geometric deviation observed under nominal conditions).

It is worth noting that both segmentation methods achieved comparable reconstruction accuracy under nominal manufacturing conditions when trained using images from both specimens. The different behavior observed in the presence of geometric anomalies therefore suggests that the two architectures differ primarily in their ability to preserve previously unseen shape deviations during reconstruction. One possible explanation is that the conventional U-Net, owing to its inductive bias toward the nominal geometries observed during training, tends to reconstruct the expected shape even in the presence of moderate deviations. In contrast, the graph-based relational representation embedded within the bottleneck of the proposed UNet-GNN may better preserve localized geometric perturbations by exploiting global structural relationships, thereby enabling the subsequent control chart to detect these deviations. These observations provide preliminary evidence supporting the hypothesis underlying this work. Nevertheless, a more comprehensive study focusing specifically on this "geometric bias" exhibited by the U-Net is ongoing, involving a larger number of anomaly types and controlled severity levels.

*4.4 Ablation study and computational efficiency*

The conventional U-Net architecture, already considered as a benchmark in the previous sections, can be regarded as a fully ablated version of the proposed UNet-GNN, in which the graph-based module is removed from the bottleneck. To further investigate the influence of the GNN depth on the geometry reconstruction performance, an additional ablation study was conducted by varying the number of GNN layers. Specifically, the original configuration with $T = 2$ layers was compared with a simplified architecture employing $T = 1$ layer and a deeper configuration with $T = 3$ layers. Table VII reports the corresponding mean accuracy and IoU values, together with their 95% confidence intervals, for the different training and inference configurations in case study 1, separately considering nodes and struts.

**TABLE VII –** Mean accuracy (top) and mean IoU (bottom) with corresponding 95% confidence intervals for the UNet-GNN with different numbers of GNN layers in the bottleneck, i.e., $T = 1, 2, 3$; reported values refer to different geometrical features (nodes and struts) of lattice structures in case study 1

| Accuracy | | | | | | | |
|---|---|---|---|---|---|---|---|
| Train | Inference | $T = 1$ | | $T = 2$ | | $T = 3$ | |
| | | Nodes | Struts | Nodes | Struts | Nodes | Struts |
| A | A | 0.987 [0.986, 0.987] | 0.980 [0.980, 0.980] | 0.987 [0.986, 0.987] | 0.980 [0.980, 0.981] | 0.987 [0.987, 0.988] | 0.981 [0.981, 0.981] |
| A | B | 0.979 [0.978, 0.979] | 0.969 [0.968, 0.971] | 0.978 [0.977, 0.979] | 0.969 [0.968, 0.971] | 0.978 [0.977, 0.979] | 0.963 [0.961, 0.966] |
| B | A | 0.976 [0.975, 0.977] | 0.975 [0.974, 0.975] | 0.977 [0.976, 0.978] | 0.976 [0.975, 0.976] | 0.978 [0.976, 0.979] | 0.976 [0.975, 0.976] |
| B | B | 0.987 [0.987, 0.988] | 0.980 [0.979, 0.981] | 0.987 [0.987, 0.988] | 0.980 [0.980, 0.981] | 0.987 [0.987, 0.988] | 0.980 [0.980, 0.981] |
| A and B | A | 0.981 [0.980, 0.982] | 0.970 [0.969, 0.970] | 0.981 [0.980, 0.982] | 0.970 [0.969, 0.971] | 0.981 [0.980, 0.982] | 0.969 [0.969, 0.970] |
| A and B | B | 0.979 [0.978, 0.980] | 0.968 [0.967, 0.969] | 0.979 [0.979, 0.980] | 0.968 [0.967, 0.969] | 0.979 [0.978, 0.980] | 0.968 [0.967, 0.969] |

| IoU | | | | | | | |
|---|---|---|---|---|---|---|---|
| Train | Inference | $T = 1$ | | $T = 2$ | | $T = 3$ | |
| | | Nodes | Struts | Nodes | Struts | Nodes | Struts |
| A | A | 0.895 [0.890, 0.899] | 0.887 [0.885, 0.889] | 0.894 [0.890, 0.898] | 0.888 [0.886, 0.890] | 0.898 [0.894, 0.902] | 0.892 [0.890, 0.894] |
| A | B | 0.827 [0.816, 0.837] | 0.834 [0.830, 0.839] | 0.824 [0.812, 0.836] | 0.834 [0.828, 0.841] | 0.822 [0.813, 0.832] | 0.801 [0.789, 0.812] |
| B | A | 0.821 [0.810, 0.832] | 0.864 [0.862, 0.867] | 0.828 [0.818, 0.838] | 0.867 [0.865, 0.869] | 0.830 [0.819, 0.842] | 0.868 [0.865, 0.872] |
| B | B | 0.903 [0.899, 0.907] | 0.893 [0.891, 0.895] | 0.903 [0.900, 0.907] | 0.893 [0.891, 0.895] | 0.903 [0.900, 0.907] | 0.893 [0.891, 0.895] |
| A and B | A | 0.848 [0.844, 0.853] | 0.834 [0.831, 0.837] | 0.850 [0.847, 0.854] | 0.836 [0.833, 0.839] | 0.848 [0.844, 0.852] | 0.833 [0.830, 0.836] |
| A and B | B | 0.844 [0.840, 0.849] | 0.836 [0.833, 0.838] | 0.847 [0.842, 0.851] | 0.836 [0.833, 0.838] | 0.846 [0.841, 0.850] | 0.837 [0.835, 0.839] |

Overall, the three configurations with different numbers of GNN layers exhibit very similar reconstruction performance. While minor differences emerge in specific conditions, there is no statistically significant improvement associated with increasing the number of GNN layers from $T = 1$ to $T = 3$. These findings suggest that the main benefits of graph augmentation arise from the introduction of relational feature processing rather than from increasing the depth of the GNN module, at least within the investigated range. From a practical perspective, a shallow GNN may therefore represent an effective compromise between reconstruction performance and architectural complexity.

Beyond reconstruction accuracy, however, the applicability of the proposed approach also depends on its computational requirements. This aspect is particularly relevant for industrial deployment, where the additional operations introduced by graph augmentation should not compromise the processing speed required for *in situ* monitoring and inspection. Accordingly, the computational efficiency of the UNet-GNN architecture examined against competitor methods was assessed.

Active contours do not require any prior training phase, but their inference relies on iterative optimization procedures that must be executed independently for each image. As a result, the processing time per layer can be long, especially when high-resolution images are considered. In contrast, deep learning-based segmentation models require an initial training phase based on annotated datasets, which can be computationally demanding in terms of both time and hardware resources. Once trained, however, inference is performed through a single forward pass of the network, resulting in significantly reduced and consistent processing times per image.

Fig. 14 shows the boxplots of per-image inference times for the three considered methods. Inference times were estimated on a Dell Precision 7920 Rack XCTO (with Intel Xeon Gold 6130, 96Gb RAM, NVIDIA Quadro P5000 GPU).

As expected, the active contour methodology exhibits significantly longer inference times, on the order of 16 – 17 s per image, due to its iterative nature. In the present study, this duration is comparable to the time required to manufacture a single layer containing only two lattice specimens. Both U-Net and UNet-GNN models unlock a much faster inference, below 1 s per image. Among the two architectures, UNet-GNN exhibited slightly lower average inference time (approximately 0.4 s per image). Although this difference may depend on architectural and implementation-specific factors, the results suggest that incorporating graph-based operations within the U-Net bottleneck does not necessarily entail an increase in inference time, at least under the computational conditions investigated in this study.

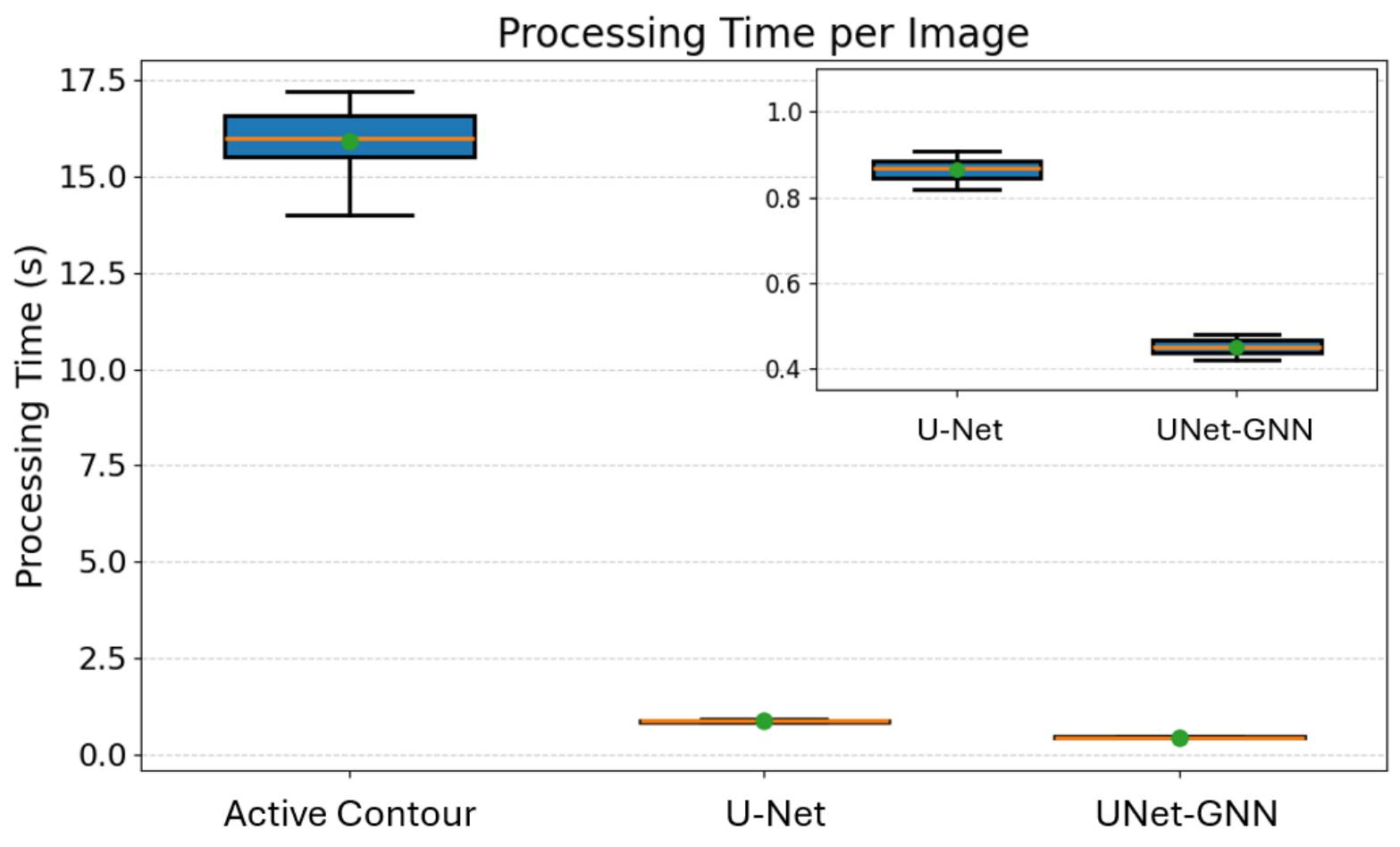

**Fig. 14.** Boxplot of the processing time per image for each considered method in the inference phase (the box in the upper-right corner is a detail of the processing time for the U-Net and UNet-GNN methods).

## V. Conclusion

This study investigated the potential of a graph-augmented segmentation approach for *in situ* inspection and geometry reconstruction in PBF-LB/M, addressing the limitations of existing methodologies related to photometric OOD generalization. By integrating a GNN bottleneck within a U-Net architecture, the resulting hybrid model enables the explicit modeling of relational dependencies among image regions in the feature space, thereby enhancing robustness to the variability inherent to industrial powder bed imaging.

The experimental results showed that the UNet-GNN model outperformed both active contour methods and standard U-Net architectures across the considered scenarios involving different geometries, materials, and experimental setups. The performance gains were more evident under heterogeneous training–testing conditions and in the presence of controlled perturbations affecting image quality, which highlights the benefits of modelling more global structural information through the GNN bottleneck. This robustness represents a key requirement for enabling *in situ* inspection as a reliable tool to support process qualification and product verification in industrially compliant implementations.

From a computational perspective, UNet-GNN maintains fast inference times comparable to standard U-Net models, while significantly outperforming iterative approaches such as active contours, thus meeting the latency constraints required for real-time monitoring.

An equally important aspect concerns the relationship between OOD generalization and anomaly detectability. Increasing robustness to undesired sources of image variability should not come at the expense of reduced sensitivity to process-induced geometric deviations. The preliminary anomaly detection results showed that a graph-augmented framework not only preserves the capability to identify actual geometric deviations but can also outperform a conventional U-Net when moderate shape distortions are introduced. The focus of the present work was the photometric OOD generalization, which has been poorly addressed in the PBF-LB/M literature so far, but an important field of future research regards the geometric OOD generalization. A more comprehensive investigation involving different geometries and a broader range of anomalies and severity levels is currently underway and will be the subject of future work. The optimization of network's hyperparameters, particularly the number $k$ of nearest neighbors considered in the GNN bottleneck is also part of future research.

Overall, the integration of graph-based learning within *in situ* monitoring and *in situ* inspection frameworks represents a promising direction toward more robust, scalable, and efficient quality assurance practices in AM.

## Declarations

The authors have no conflict of interest to declare that are relevant to the content of this article.

## Acknowledgements

The authors would like to acknowledge Prof. Bianca Maria Colosimo for coordinating the research activities that contributed to this work. The PhD scholarship of Raimondo Stefano is funded by GE Avio.

## Authors' contribution

S. Raimondo: Data curation, methodology, verification, writing, and revision. M. Bugatti: Data curation, methodology, verification, writing, and revision. M. Grasso: Conceptualization, methodology, verification, supervision, writing, and revision.

## Appendix

This section complements results presented in Fig. 10. In particular, Fig. A1 shows the 95% confidence intervals for the mean IoU computed in regions corresponding respectively to nodes and struts of the trabecular lattice structures in case study 1. Fig. A2 and Fig. A3 show the 95% confidence intervals for the mean accuracy and the mean IoU computed in correspondence of two different geometrical features of the gyroid lattice structures in case study 2, namely a region where separate walls are present, and a region where walls approach and merge together.

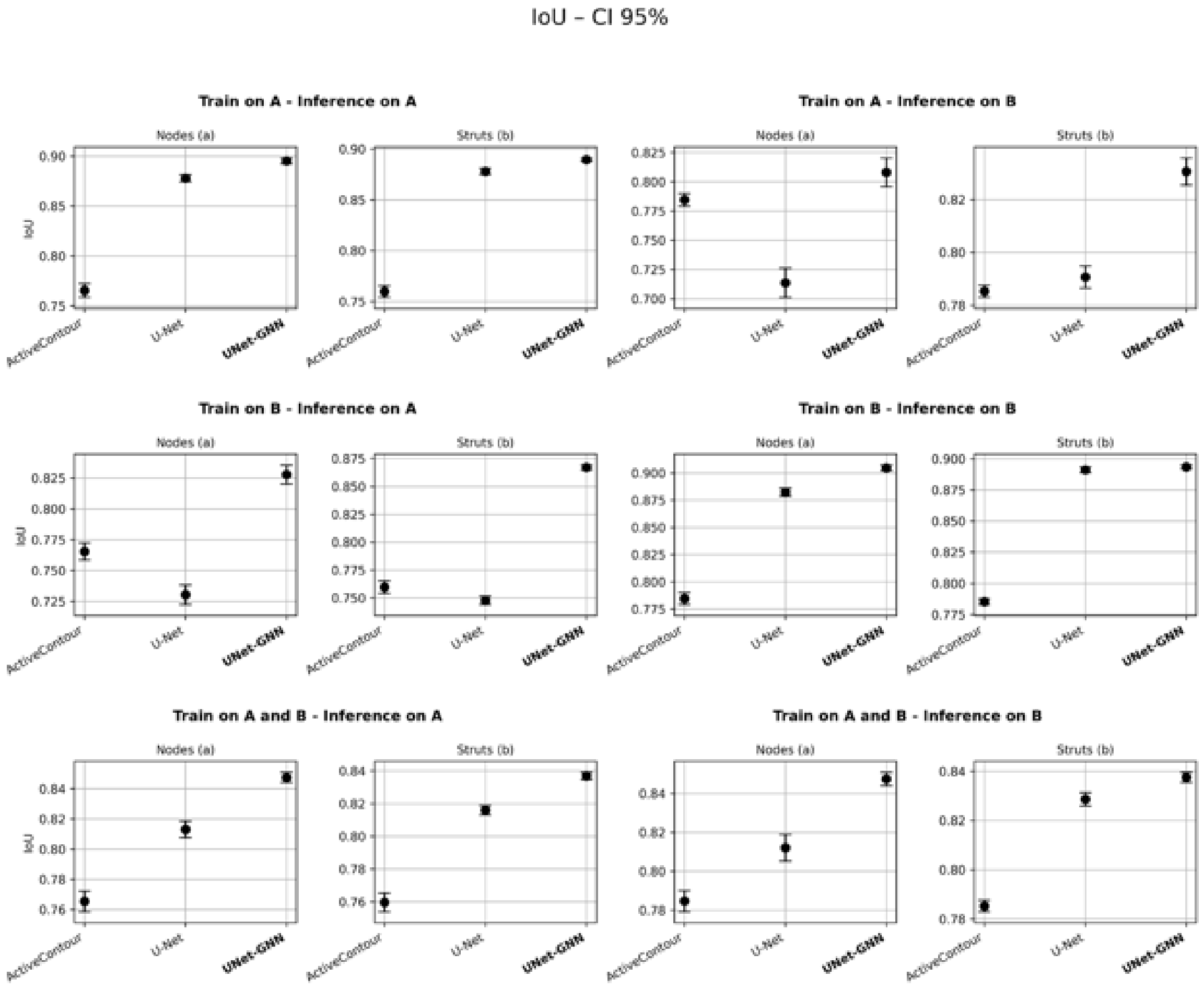

**Fig. A1.** 95% confidence intervals for the mean IoU computed over regions a (nodes) and b (struts) of lattice structures in case study 1 for all compared segmentation methods. The y-axis range is independently scaled for each subplot to emphasize inter-method differences.

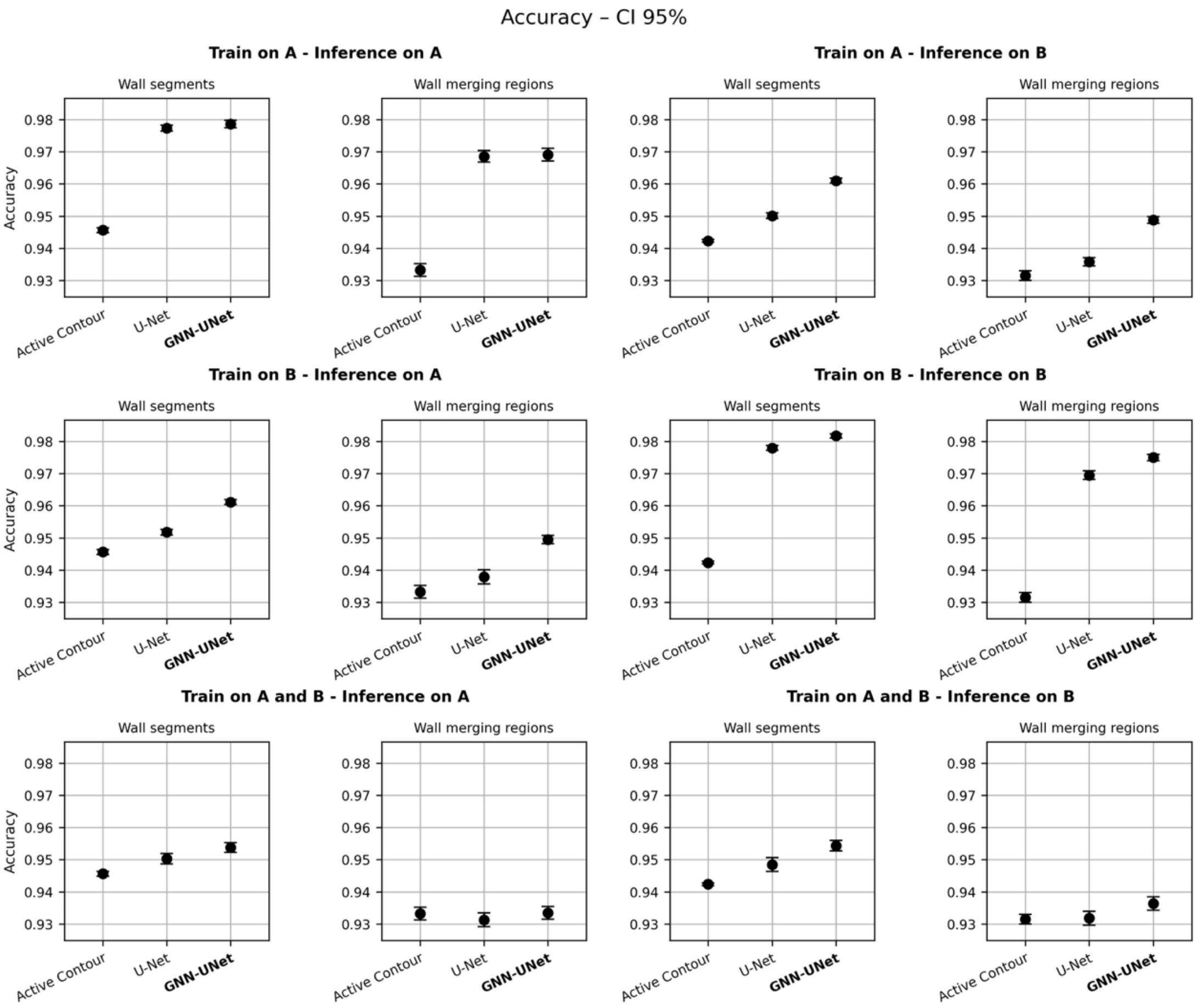

**Fig. A2.** 95% confidence intervals for the mean accuracy computed over regions a (separate wall segments) and b (wall-merging regions) of lattice structures in case study 2 for all compared segmentation methods. The y-axis range is independently scaled for each subplot to emphasize inter-method differences.

Fig. A1confirms the statistically significant differences observed in Fig. 10 for case study 1. The results presented in Figs. A2 and A3 confirm the findings obtained for case study 1, too. The UNet-GNN generally achieves the highest accuracy and IoU across the investigated training and inference configurations, for both separate wall segments and wall-merging regions of the gyroid lattice structure. The benefits of graph augmentation are particularly evident under cross-dataset inference conditions, whereas smaller differences between UNet-GNN and conventional U-Net are observed when training and inference conditions are homogeneous or when joint training is adopted. The improvements are observed for both geometric features, although their magnitude varies depending on the training configuration and performance metric.

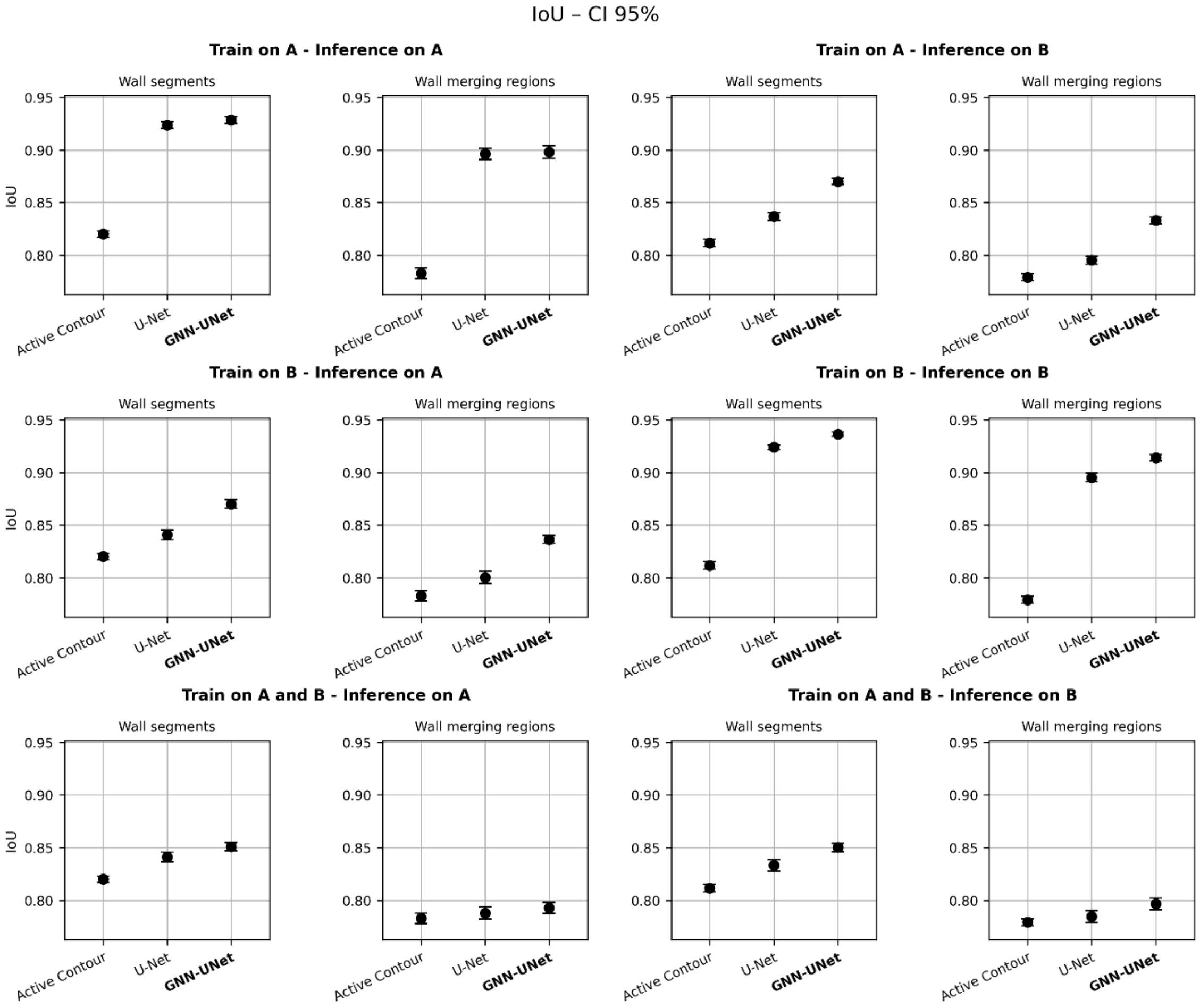

**Fig. A3.** 95% confidence intervals for the mean IoU computed over regions a (separate wall segments) and b (wall-merging regions) of lattice structures in case study 2 for all compared segmentation methods. The y-axis range is independently scaled for each subplot to emphasize inter-method differences.